\documentclass{article}

    \PassOptionsToPackage{numbers, compress}{natbib}
 \usepackage[preprint]{neurips_2026}

\usepackage[utf8]{inputenc} 
\usepackage[T1]{fontenc}    
\usepackage{hyperref}       
\usepackage{url}            
\usepackage{booktabs}       
\usepackage{amsfonts}       
\usepackage{nicefrac}       
\usepackage{microtype}      
\usepackage{xcolor}         

\usepackage{amsmath}
\usepackage{graphicx}

\title{Model-Free Neural Filtering: A Comparison with Classical Filters in Nonlinear Systems}

\author{%
  Zhuochen Liu \\
  University of Southern California \\
  Los Angeles, CA 90089 \\
  \texttt{liuzhuoc@usc.edu} \\
  \AND
  Hans Walker \\
  University of Southern California \\
  Los Angeles, CA 90089 \\
  \texttt{hbwalker@usc.edu} \\
  \And
  Rahul Jain \\
  University of Southern California \\
  Los Angeles, CA 90089 \\
  \texttt{rahul.jain@usc.edu} \\
}

\begin{document}

\maketitle

\begin{abstract}
Neural network models are increasingly used for state estimation in control and decision-making, yet it remains unclear to what extent they behave as principled filters in nonlinear dynamical systems. Unlike classical filters, which rely on explicit dynamics and noise models, neural estimators can be trained purely from data. We present a systematic comparison between model-free neural estimators and classical filtering methods across multiple nonlinear scenarios. On the neural side, we evaluate Transformer-based models, recurrent neural networks, and state-space models; on the classical side, we compare against particle filters and nonlinear Kalman filters. Results show that structured state-space models (SSMs), in particular Mamba and Mamba-2, are consistently strong among neural estimators. They approach strong classical filters in several nonlinear systems and outperform weaker classical baselines without access to system models, while the evaluated neural implementations achieve substantially higher inference throughput on the tested hardware. Accurate model-based filters can still dominate when their assumptions are well matched. We attribute the relative strength of SSMs to filtering-aligned inductive bias: recursive latent-state updates make them structurally closer to classical filters under fixed parameter budgets, finite data, and long-horizon evaluation.

\end{abstract}

\section{Introduction}
\label{sec:introduction}

Control systems operate under uncertainty from process noise, sensor noise, and partial observability. In robotics, autonomous systems, and cyber-physical systems, the dynamics are often nonlinear and the system state is only partially observed, making accurate state estimation essential for control.

Filtering provides a principled framework for this problem. Classical filters model the dynamics and observation process explicitly, then recursively estimate latent states through probabilistic inference. Linear-Gaussian systems admit closed-form solutions, while nonlinear systems are typically handled by approximate filters. These methods remain widely used because they are interpretable and effective when accurate models are available.

However, classical filters can degrade in highly nonlinear systems, under model mismatch, or when dynamics and observation models are only partially known \citep{judd2003nonlinear, harlim2008filtering, berry2013adaptive, chauchat2021robust}. These limitations have motivated neural approaches that learn state estimators directly from data, often without explicit system dynamics \citep{al2019deep, zhang2019real, ghosh2024danse, dahal2024robuststatenet}. Despite promising results, their filtering behavior is not yet well characterized, especially in nonlinear systems and long-horizon evaluation where errors can accumulate.

A central distinction is informational: classical filters use explicit system and noise models, while neural estimators are often trained purely from data, without access to equations or noise distributions. This raises the question: \textit{to what extent can model-free neural state estimators match model-based classical filters in nonlinear systems where accurate modeling may be difficult or unavailable?}

Recent work has analyzed connections between neural architectures and classical filters under specific assumptions on system structure or noise \citep{gu2017dynamic, gu2021combining, monga2021algorithm}. These studies provide useful insights but remain limited in scope and do not fully characterize practical performance across nonlinear systems, neural architectures, and long-horizon regimes.

\textbf{Main Contributions.}
We study these questions through controlled experiments across multiple nonlinear dynamical scenarios. Our contributions are:
\begin{enumerate}
    \item We provide a unified empirical comparison of classical filters and model-free neural estimators, including recurrent networks, Transformer-based models, and structured state-space models, across accuracy, robustness, data efficiency, stability, outlier behavior, and inference throughput.

    \item We show that, under a fixed parameter budget and supervised training protocol, Mamba and Mamba-2 are the strongest neural estimators in our benchmark. They approach strong classical filters in several nonlinear systems without using system equations or noise models, while model-based filters remain superior when their assumptions are well matched.

    \item We analyze why structured state-space models are naturally aligned with recursive filtering and how this bias can benefit learned nonlinear state estimation under finite data and long horizons.

    \item We release the experimental code to support reproducibility and facilitate further study of model-free neural filtering: \url{https://anonymous.4open.science/r/neural-state-estimator-77B4/}.
\end{enumerate}

\section{Related Work}
\label{sec:related_work}

\textbf{Classical filtering and neural filtering methods.}
State estimation in dynamical systems is commonly formulated as Bayesian filtering, with the Kalman filter and nonlinear extensions serving as standard tools in control and estimation \citep{kalman1960}. These methods rely on explicit dynamics, observation models, and noise assumptions. When these assumptions are inaccurate, performance can degrade, motivating neural components in filtering pipelines. Prior work has learned correction terms, gain functions, dynamics components, or observation models within Kalman-style recursive estimators \citep{shaj2021action, chen2021dynanet, revach2022kalmannet, choi2023split}, while other approaches learn filtering updates directly from data without strict linear-Gaussian assumptions \citep{lim2020recurrent, yan2024explainable}. Broader reviews summarize the assumptions and trade-offs of neural and hybrid filtering methods \citep{jin2021new, kim2022review, feng2023review}.

\textbf{Neural architectures for state estimation.}
General-purpose neural architectures have also been studied for state estimation. Early empirical work compared recurrent networks with Kalman filters on tracking tasks and reported comparable performance under specific conditions \citep{chenna2004state, gao2019long}. More recently, Transformer-based models have been explored because they can use long observation histories. In linear dynamical systems, a Transformer layer can be constructed to approximate the Kalman filter update, formally connecting self-attention and classical filtering in a restricted setting \citep{goel2024can}. Empirical studies further show that Transformers trained from data can perform well in both linear and nonlinear systems and can match optimal Kalman filtering in linear cases with sufficient data \citep{du2023can}.

\textbf{Beyond linear and restricted filtering settings.}
Theoretical understanding beyond linear and Markovian systems remains limited. Recent continuous-time filtering work studies Transformer-like architectures for conditionally Gaussian signal processes, showing approximation of conditional laws for broad nonlinear and non-Markovian systems under noisy observations \citep{horvath2025transformerssolvenonlinearnonmarkovian}. In parallel, structured sequence models based on state-space formulations have emerged as efficient alternatives to attention \citep{gu2024mamba, dao2024transformers}. Recent neural Kalman filtering frameworks also use modern state-space backbones such as Mamba for efficient real-time estimation \citep{sun2025towards}. Our work complements these directions by comparing recurrent, Transformer-based, structured state-space, and classical filters under one protocol across multiple nonlinear systems.

Our focus differs from hybrid learned-filtering methods that preserve Kalman-style or particle-filter structure while learning selected components. Such methods are important when partial model structure is available. Here, we isolate fully model-free sequence estimators that map causal observation-control histories directly to state estimates, leaving systematic comparison with hybrid learned filters to future work.

\section{Preliminaries}
\label{sec:preliminaries}

We consider discrete-time nonlinear state-space systems
\begin{equation}
\label{eq:nonlinear_system}
x_{t+1}=f(x_t,u_t)+w_t,\qquad y_t=h(x_t)+v_t,
\end{equation}
where $x_t$ is the latent state, $u_t$ the control input, $y_t$ the observation, and $w_t,v_t$ process and observation noise. The filtering objective is to estimate
\begin{equation}
\label{eq:filter}
p(x_t\mid y_{1:t},u_{1:t-1}).
\end{equation}
For nonlinear or non-Gaussian systems, this posterior is generally intractable, motivating approximate filters.

We compare neural estimators with four classical nonlinear filters. The \textbf{Extended Kalman Filter (EKF)} applies Kalman filtering to a locally linearized system. The \textbf{Unscented Kalman Filter (UKF)} propagates sigma points through nonlinear dynamics to approximate posterior moments. The \textbf{Ensemble Kalman Filter (EnKF)} uses ensemble covariance estimates. The \textbf{Particle Filter (PF)} represents the posterior with weighted particles and can handle non-Gaussian distributions, but may suffer from particle degeneracy. Together, these methods provide model-based baselines with different efficiency, expressiveness, and robustness trade-offs. PF and EnKF can be sensitive to particle count, ensemble size, resampling, inflation, and rejuvenation; we use fixed representative configurations for a unified benchmark rather than exhaustive compute-matched tuning.

\section{Methodology}
\label{sec:methodology}

We evaluate neural estimators and classical filters under one protocol. All methods use the same observation sequences, control inputs, horizons, and metrics. Classical filters access the dynamics, observation functions, and noise parameters specified by each benchmark. Neural estimators are trained only from data and do not use the system equations or noise distributions. Thus, the comparison is not equal-information; it measures how closely model-free neural estimators can approach strong model-based filters.

\subsection{Neural Estimators and Training}

We evaluate six neural sequence models: GRU \citep{cho2014learning}, LSTM \citep{hochreiter1997long}, GPT-2 \citep{radford2019language}, Filterformer \citep{horvath2025transformerssolvenonlinearnonmarkovian}, Mamba \citep{gu2024mamba}, and Mamba-2 \citep{dao2024transformers}. All models have approximately $100{,}000$ parameters, so our comparison targets a fixed-budget regime rather than a universal ranking across scales. For GPT-2, we replace learned absolute positional embeddings with Attention with Linear Biases (ALiBi) \citep{press2021train} to support extrapolation beyond the training horizon. Full hyperparameters appear in Appendix Table~\ref{tab:hparams}.

At timestep $t$, a neural estimator receives only causal information and predicts
\begin{equation}
\hat{x}_t = F_\theta(z_{1:t}),
\end{equation}
where $z_t=y_t$ for systems without controls and $z_t=[y_t,u_{t-1}]$ for systems with controls. Inputs are noisy observations, while supervised targets are clean simulator states.

Thus, neural estimators solve supervised causal point estimation rather than full posterior filtering. Our comparison to Bayesian filters therefore focuses on accuracy, stability, robustness, and runtime, not uncertainty calibration.

The default training setting uses a data-to-parameter ratio of approximately $20{:}1$. With about $100{,}000$ parameters, this corresponds to $20{,}000$ trajectories of length $100$, or $2{,}000{,}000$ observed timesteps. We also train under $10{:}1$, $5{:}1$, and $1{:}1$ ratios with the same parameter budget. For each scenario, we generate three training datasets and train five random initializations per dataset, yielding $15$ trained instances per neural architecture.

Training samples are random length-$100$ sub-trajectories extracted from longer trajectories of length $500$. Each training dataset has an independent validation set of $8{,}000$ trajectories for early stopping. Because chunking samples interior and endpoint timesteps with different frequencies, we weight each timestep loss by the inverse of its sampling frequency.

\subsection{Evaluation Protocol}

Each scenario uses three independent evaluation datasets with $8{,}000$ trajectories each. The default evaluation horizon is $500$. Planar Quadrotor uses horizon $200$ because, without stabilizing control, longer free-fall trajectories become uninformative. Training on length-$100$ chunks and evaluating on longer horizons tests whether neural models learn stable filtering rather than short-horizon interpolation.

\subsection{Evaluation Metrics}

We report RMSE over time, aggregate RMSE, NRMSE, MAE, MedAE, area under the RMSE-time curve (AUC), drift ratio, outlier ratio, and inference throughput.

To measure long-horizon error accumulation, we compute a drift ratio from the RMSE-time curve. Let $T$ be the effective trajectory length and $\mathrm{RMSE}_t$ the per-timestep RMSE. We divide the trajectory into quarters and compare the mean RMSE in the final quarter to that in the first quarter. With $q=\max(1,\lfloor T/4 \rfloor)$,
\begin{equation}
\mathrm{drift\_ratio}
=
\frac{
\frac{1}{q}\sum_{t=T-q+1}^{T}\mathrm{RMSE}_t
}{
\frac{1}{q}\sum_{t=1}^{q}\mathrm{RMSE}_t
}.
\end{equation}

To prevent rare catastrophic failures from dominating aggregate metrics, we detect trajectory-level outliers on log-MSE values using a robust \(z\)-score based on the median absolute deviation (MAD). Let \(M_n\) be the trajectory-level MSE for trajectory \(n\), and define
\[
\begin{aligned}
u_n &= \log M_n,\\
\tilde{u} &= \operatorname{median}\{u_1,\dots,u_N\},\\
\operatorname{MAD} &= \operatorname{median}\{|u_1-\tilde{u}|,\dots,|u_N-\tilde{u}|\}.
\end{aligned}
\]
The robust \(z\)-score is
\begin{equation}
z_n
=
\frac{|u_n-\tilde{u}|}{1.4826\,\operatorname{MAD}}.
\end{equation}
A trajectory is an outlier if \(z_n>6\). We report the outlier ratio over all trajectories. Main RMSE and NRMSE exclude detected outliers, while full metrics with outlier statistics are reported in the Appendix.

Inference speed is measured with batch size one on both GPU and CPU. GPU measurements use an NVIDIA GeForce RTX 4090, and CPU measurements use an Intel Core Ultra 9 Processor 285K. Timing excludes data generation and is performed after warm-up iterations.

\section{Experiment Setup}
\label{sec:experiment_setup}

We evaluate all methods on five nonlinear state estimation scenarios covering different dynamics, observation structures, state dimensions, and filtering difficulties.

\subsection{Scenarios}

\textbf{Ballistic re-entry.}
We use a simplified ballistic re-entry model in which an object falls under gravity through an atmosphere with quadratic drag and exponential density. The state is $\mathbf{x}(t)=[a(t),v(t),c(t)]$, where $a(t)$ is altitude, $v(t)$ downward velocity, and $c(t)$ a drag coefficient. The dynamics are
\begin{equation}
\begin{aligned}
\dot a(t) &= -v(t),\\
\dot v(t) &= g - c(t)v(t)^2E(a(t)),\\
\dot c(t) &= 0,
\end{aligned}
\end{equation}
where $g>0$ and
\[
E(a)=\min\{\exp(-ka),1\}
\]
models the atmosphere. At time $t_k$, an observer measures range
\begin{equation}
y_k=\sqrt{r_0^2 + (a(t_k)-a_{\rm ref})^2} + \nu_k,
\end{equation}
where $r_0$ is a fixed horizontal offset, $a_{\rm ref}$ is a reference altitude, and $\nu_k$ is measurement noise. The task is challenging because the observation is nonlinear and drag depends strongly on altitude and velocity.

\textbf{Bearings-only tracking.}
We consider two-dimensional single-sensor bearings-only target tracking with a constant-velocity motion model \citep{lascala_morelande2008_bot}. The state is $\mathbf{x}_k=[p_{x,k},p_{y,k},v_{x,k},v_{y,k}]$ with dynamics
\begin{equation}
\mathbf{x}_{k+1} =
\begin{bmatrix}
1 & 0 & T & 0\\
0 & 1 & 0 & T\\
0 & 0 & 1 & 0\\
0 & 0 & 0 & 1
\end{bmatrix}\mathbf{x}_k+\mathbf{w}_k,
\end{equation}
where $T$ is the sampling period. The observation is the bearing from the origin,
\begin{equation}
y_k=\mathrm{wrap}\!\left(\arctan2(p_{y,k},p_{x,k})\right)+\nu_k,
\end{equation}
with $\mathrm{wrap}(\cdot)$ mapping angles to $(-\pi,\pi]$. The problem is ill-conditioned because range is unobserved and must be inferred from bearing history.

\textbf{Lorenz 96.}
The Lorenz 96 system is a standard chaotic benchmark for filtering and data assimilation \citep{lorenz96_original}:
\begin{equation}
\frac{dX_j}{dt}
=
(X_{j+1}-X_{j-2})X_{j-1}-X_j+F,
\qquad j=1,\dots,J,
\end{equation}
with cyclic boundary conditions $X_{-1}=X_{J-1}$, $X_0=X_J$, and $X_{J+1}=X_1$. The forcing parameter $F$ controls the regime, with values near $F=8$ producing chaos. This scenario tests long-horizon stability under nonlinear evolution.

\textbf{N-link pendulum.}
The $N$-link pendulum has $N$ rigid links of lengths $R_1,\dots,R_N$ and endpoint masses $m_1,\dots,m_N$. Let $\theta_j$ be the angle of link $j$ from vertical and $\omega_j=\dot{\theta}_j$. The Lagrangian of the $i$th mass is
\begin{equation}
\begin{aligned}
\mathcal{L}_i(\theta,\omega)
&=
\frac12 m_i
\sum_{j=1}^i\sum_{k=1}^i
R_jR_k\omega_j\omega_k\cos(\theta_j-\theta_k)\\
&\quad
+
m_i g\sum_{j=1}^i R_j\cos(\theta_j).
\end{aligned}
\end{equation}
The total Lagrangian is
\[
\mathcal{L}(\theta,\omega)=\sum_{i=1}^N\mathcal{L}_i(\theta,\omega),
\]
and the Euler-Lagrange equations yield strongly coupled nonlinear dynamics. The task is challenging because the dynamics are highly coupled and sensitive to state errors.

\textbf{Planar Quadrotor.}
We consider a planar quadrotor constrained to a vertical plane \citep{singh2023robust}. The state is
\[
\mathbf{x}_t=[x_t,z_t,\phi_t,\dot{x}_t,\dot{z}_t,\dot{\phi}_t],
\]
and the dynamics are
\begin{equation}
\begin{aligned}
\mathbf{x}_{t+1}
&=
\begin{bmatrix}
x_t + (\dot{x}_t \cos\phi_t - \dot{z}_t \sin\phi_t)\tau \\
z_t + (\dot{x}_t \sin\phi_t + \dot{z}_t \cos\phi_t)\tau \\
\phi_t + \dot{\phi}_t \tau \\
\dot{x}_t + (\dot{z}_t \dot{\phi}_t - g \sin\phi_t)\tau \\
\dot{z}_t+(-\dot{x}_t\dot{\phi}_t-g\cos\phi_t+(u_{0,t}+u_{1,t})/m)\tau \\
(u_{0,t}-u_{1,t})l\tau/J
\end{bmatrix}
+
\mathbf{w}_t,\\
\mathbf{y}_t
&=
\mathbf{C}\mathbf{x}_t+\mathbf{v}_t,
\end{aligned}
\end{equation}
where $\mathbf{w}_t$ and $\mathbf{v}_t$ are process and observation noise, and $\mathbf{C}\in\mathbb{R}^{3\times6}$ is a fixed randomly sampled observation matrix. This scenario tests estimation under nonlinear controlled dynamics and partial observation.

\section{Results}
\label{sec:results}

We evaluate neural estimators and classical filters across the five scenarios using the metrics in Section~\ref{sec:methodology}. Unless noted otherwise, results are summarized over independent datasets and random initializations using medians with interquartile ranges. Detailed per-scenario metrics appear in Appendix Tables~\ref{tab:ballistic_reentry_extended}--\ref{tab:planar_quadrotor_extended}.

\subsection{Accuracy Across Scenarios}

Table~\ref{tab:accuracy_main} reports the median RMSE and NRMSE for each method in each scenario. RMSE preserves the absolute estimation error scale within each benchmark, while NRMSE provides a normalized accuracy measure that accounts for scenario-specific state scale. This table focuses on scenario-level accuracy; aggregate NRMSE summaries and relative performance comparisons are reported separately in the Appendix and in Table~\ref{tab:relative_score}. Full median and interquartile-range statistics for RMSE, NRMSE, MAE, MedAE, AUC, outlier ratio, and drift ratio are reported in Appendix Tables~\ref{tab:ballistic_reentry_extended}--\ref{tab:planar_quadrotor_extended}.

\begin{table}[htbp]
  \centering
  \caption{Median RMSE and NRMSE across scenarios. Lower is better. Bold indicates the best method within each method class. Full median $\pm$ IQR statistics are reported in the Appendix.}
  \label{tab:accuracy_main}
  \footnotesize
  \setlength{\tabcolsep}{1.3pt}
  \begin{tabular}{lcccccccccc}
    \toprule
    & \multicolumn{2}{c}{Ballistic}
    & \multicolumn{2}{c}{Bearings}
    & \multicolumn{2}{c}{L96}
    & \multicolumn{2}{c}{N-Link}
    & \multicolumn{2}{c}{P. Quad.} \\
    \cmidrule(lr){2-3}
    \cmidrule(lr){4-5}
    \cmidrule(lr){6-7}
    \cmidrule(lr){8-9}
    \cmidrule(lr){10-11}
    Method
    & RMSE & NRMSE
    & RMSE & NRMSE
    & RMSE & NRMSE
    & RMSE & NRMSE
    & RMSE & NRMSE \\
    \midrule
    \multicolumn{11}{l}{\textit{Classical Filters}} \\
    EKF
    & 5.289 & 1.571
    & 308.624 & 1.005
    & 0.049 & 0.014
    & \textbf{0.080} & \textbf{0.012}
    & \textbf{0.287} & \textbf{0.009} \\
    UKF
    & \textbf{0.626} & \textbf{0.607}
    & \textbf{180.946} & \textbf{0.647}
    & \textbf{0.048} & \textbf{0.013}
    & 0.081 & \textbf{0.012}
    & 8.042 & 0.073 \\
    EnKF
    & 0.821 & 0.878
    & $\infty$ & $\infty$
    & 0.054 & 0.015
    & 0.092 & 0.014
    & 9.960 & 0.092 \\
    PF
    & 0.657 & 0.648
    & 362.554 & 1.429
    & 0.065 & 0.018
    & 0.086 & 0.550
    & 128.009 & 0.957 \\
    \midrule
    \multicolumn{11}{l}{\textit{Neural Models}} \\
    GRU
    & 0.676 & 0.629
    & 191.896 & 0.637
    & 0.127 & 0.036
    & 0.399 & 0.063
    & 13.023 & 0.224 \\
    LSTM
    & 0.715 & 0.699
    & 187.855 & 0.616
    & 0.119 & 0.034
    & 0.356 & 0.058
    & 11.444 & 0.219 \\
    GPT-2
    & 0.759 & 0.696
    & 225.685 & 0.699
    & 0.099 & 0.028
    & 0.233 & 0.037
    & 16.420 & 0.262 \\
    Filterformer
    & 0.793 & 0.717
    & 224.502 & 0.744
    & 0.732 & 0.202
    & 1.749 & 0.219
    & 14.345 & 0.254 \\
    Mamba
    & 0.654 & 0.613
    & \textbf{168.081} & \textbf{0.588}
    & \textbf{0.052} & \textbf{0.015}
    & 0.093 & 0.017
    & 7.395 & 0.183 \\
    Mamba-2
    & \textbf{0.631} & \textbf{0.593}
    & 177.842 & 0.611
    & \textbf{0.052} & \textbf{0.015}
    & \textbf{0.092} & \textbf{0.015}
    & \textbf{5.702} & \textbf{0.175} \\
    \bottomrule
  \end{tabular}
\end{table}

The scenario-level results show that structured state-space models are consistently strong among neural estimators, but the relative gaps vary across systems. Mamba-2 achieves the best neural performance in Ballistic Re-entry, N-link Pendulum, and Planar Quadrotor, while Mamba performs best in Bearings-only Tracking. In Lorenz 96, Mamba and Mamba-2 are close to the strongest classical filters, and in Bearings-only Tracking, Mamba is competitive with UKF. In Planar Quadrotor, however, EKF is substantially better than all neural estimators, showing that accurate model-based filtering can remain decisively superior when its assumptions are well matched.

Presenting RMSE and NRMSE together gives two complementary views of accuracy. RMSE shows absolute estimation error within each benchmark, while NRMSE normalizes for scenario-specific scale. This scenario-level comparison provides the basis for the following relative-performance analysis in Table~\ref{tab:relative_score}.

\subsection{Relative Quantitative Comparison}

To make performance gaps explicit, Table~\ref{tab:relative_score} reports RMSE normalized to Mamba-2:
\begin{equation}
\label{eq:relative_score}
\mathrm{score}(m,s)
=
100
\times
\frac{\mathrm{RMSE}(\mathrm{Mamba\mbox{-}2},s)}
{\mathrm{RMSE}(m,s)},
\end{equation}
where $m$ denotes the method and $s$ the scenario. A score of $100$ matches Mamba-2, above $100$ indicates lower RMSE, and below $100$ indicates higher RMSE.

\begin{table}[htbp]
  \centering
  \caption{Relative RMSE score normalized to Mamba-2. Higher is better. A score of $100$ matches Mamba-2.}
  \label{tab:relative_score}
  \footnotesize
  \begin{tabular}{lccccc}
    \toprule
    Method & Ballistic & Bearings & L96 & N-Link & P. Quad. \\
    \midrule
    \multicolumn{6}{l}{\textit{Classical Filters}} \\
    EKF & 11.9 & 57.6 & 107.0 & 115.5 & 1988.8 \\
    UKF & 100.9 & 98.3 & 109.4 & 113.5 & 70.9 \\
    EnKF & 76.9 & 0.0 & 97.2 & 99.5 & 57.2 \\
    PF & 96.0 & 49.1 & 80.6 & 106.1 & 4.5 \\
    \midrule
    \multicolumn{6}{l}{\textit{Neural Models}} \\
    GRU & 93.4 & 92.7 & 41.2 & 23.0 & 43.8 \\
    LSTM & 88.4 & 94.7 & 44.0 & 25.8 & 49.8 \\
    GPT-2 & 83.2 & 78.8 & 52.8 & 39.5 & 34.7 \\
    Filterformer & 79.6 & 79.2 & 7.1 & 5.2 & 39.7 \\
    Mamba & 96.6 & 105.8 & 100.2 & 98.4 & 77.1 \\
    Mamba-2 & 100.0 & 100.0 & 100.0 & 100.0 & 100.0 \\
    \bottomrule
  \end{tabular}
\end{table}

Mamba and Mamba-2 achieve the strongest overall accuracy among neural estimators across scenarios. The largest neural gap appears in N-link Pendulum, where GRU and LSTM score only $23.0$ and $25.8$ relative to Mamba-2. In Lorenz 96, Mamba and Mamba-2 nearly match the strongest classical filters; in Bearings-only tracking, Mamba slightly outperforms Mamba-2 and is competitive with UKF. Planar Quadrotor shows the clearest advantage for model-based filtering, with EKF far ahead of all neural models.

\subsection{Stability and Outlier Behavior}

Appendix Table~\ref{tab:stability_summary_app} reports per-scenario outlier and drift ratios. These metrics expose failures not captured by typical RMSE: a method can have competitive outlier-filtered RMSE but much worse Full RMSE or drift due to rare catastrophic trajectories.

The results show that accuracy and long-horizon robustness are not identical. PF is competitive in typical RMSE for some scenarios but has large drift or outlier rates in N-link Pendulum and Planar Quadrotor. Neural models are also scenario-dependent: Mamba and Mamba-2 often combine strong accuracy with favorable long-horizon trends, but no neural architecture dominates every stability metric.

\subsection{Data Efficiency}

The main comparison uses a data-to-parameter ratio of approximately $20{:}1$. We additionally train all neural estimators under $10{:}1$, $5{:}1$, and $1{:}1$ ratios with fixed parameter budgets. Table~\ref{tab:data_ratio_main} summarizes median NRMSE across scenarios, while per-scenario data-ratio results and comparison are reported in Appendix Figures~\ref{fig:data_ratio_ballistic}--\ref{fig:data_ratio_quadrotor}.

\begin{table}[htbp]
  \centering
  \caption{Effect of data-to-parameter ratio on neural estimator performance. Entries report median NRMSE across scenarios. Lower is better. The $20{:}1$ setting is the default setting used in the main comparison.}
  \label{tab:data_ratio_main}
  \footnotesize
  \begin{tabular}{lcccc}
    \toprule
    Method & $20{:}1$ & $10{:}1$ & $5{:}1$ & $1{:}1$ \\
    \midrule
    GRU          & 0.224 & 0.251 & 0.255 & 0.307 \\
    LSTM         & 0.219 & 0.254 & 0.255 & 0.324 \\
    GPT-2        & 0.262 & 0.285 & 0.302 & 0.322 \\
    Filterformer & 0.254 & 0.290 & 0.294 & 0.339 \\
    Mamba        & 0.183 & 0.229 & 0.213 & 0.306 \\
    Mamba-2      & \textbf{0.175} & \textbf{0.214} & \textbf{0.210} & \textbf{0.292} \\
    \bottomrule
  \end{tabular}
\end{table}

Mamba-2 achieves the best neural accuracy across all data-to-parameter ratios, with Mamba usually ranking second. Performance generally worsens as data availability decreases relative to the default $20{:}1$ setting, but structured SSMs remain strongest beyond the high-data regime.

\subsection{Robustness to Observation Noise}

We evaluate robustness to observation-noise levels beyond training and report per-scenario noise-sensitivity curves in Appendix Figures~\ref{fig:noise_ballistic}--\ref{fig:noise_quadrotor}. These experiments test whether neural estimators remain stable outside the training noise regime.

\subsection{Runtime Performance}

Table~\ref{tab:runtime_summary} reports average inference throughput in iterations per second across scenarios. We include GPU and CPU columns because GPU results can benefit from optimized neural-network kernels, while CPU speed better reflects deployment settings without high-end accelerators. Full per-scenario runtime results are reported in Appendix Table~\ref{tab:runtime_full_app}.

\begin{table}[htbp]
  \centering
  \caption{Average inference throughput across scenarios in iterations per second. Higher is better. GPU and CPU results are reported separately to distinguish accelerator throughput from deployment-oriented CPU performance.}
  \label{tab:runtime_summary}
  \footnotesize
  \begin{tabular}{lccc}
    \toprule
    Method & CPU Avg. & GPU Avg. & CPU/GPU Ratio \\
    \midrule
    \multicolumn{4}{l}{\textit{Classical Filters}} \\
    EKF & 2888.3 & 864.8 & 3.34 \\
    UKF & 2710.0 & 967.0 & 2.80 \\
    EnKF & 5754.8 & 1695.8 & 3.39 \\
    PF & 1372.4 & 1654.9 & 0.83 \\
    \midrule
    \multicolumn{4}{l}{\textit{Neural Models}} \\
    GRU & 31611.8 & 1156492.2 & 0.03 \\
    LSTM & 83198.7 & 1009852.5 & 0.08 \\
    GPT-2 & 171082.8 & 425297.7 & 0.40 \\
    Filterformer & 100138.0 & 2806349.0 & 0.04 \\
    Mamba & N/A & 1376493.7 & N/A \\
    Mamba-2 & N/A & 396025.5 & N/A \\
    \bottomrule
  \end{tabular}
\end{table}

Neural models achieve high throughput on the tested hardware. CPU measurements are available for GRU, LSTM, GPT-2, and Filterformer because they use standard PyTorch operations. The official \texttt{mamba\_ssm} implementation used for Mamba and Mamba-2 relies on CUDA/Triton-backed kernels \citep{tillet2019triton} for the tested forward paths, so their CPU entries are N/A rather than failed or artificially modified CPU baselines.

\section{Architectural Inductive Bias Analysis}
\label{sec:architectural_bias_analysis}

The empirical results suggest that structured state-space models provide a useful bias for learned nonlinear filtering. The goal here is not to prove that SSMs are optimal filters, but to explain why their computation is better aligned with recursive state estimation than generic recurrence or causal attention under finite data and long horizons.

\subsection{Filtering as Recursive State Estimation}

A causal filter maintains a latent summary of past information and updates it recursively:
\begin{equation}
\label{eq:belief_recursion}
b_t = \Phi(b_{t-1}, y_t, u_{t-1}),
\qquad
\hat{x}_t = \Psi(b_t).
\end{equation}
In the linear-Gaussian case, the Kalman filter instantiates this recursive compression exactly. For example, in a strictly causal linear setting, the state estimate can be written as
\begin{equation}
\label{eq:kalman_recursion}
\hat{x}_t = (A-LC)\hat{x}_{t-1} + L y_{t-1},
\end{equation}
for suitable system matrices $A,C$ and Kalman gain $L$. Since Transformers can represent Kalman filtering-like computations in linear systems \citep{goel2024can}, the issue is not expressivity alone, but which architecture is biased toward learning stable recursive filtering from finite data.

\subsection{Architectural Computation Patterns}

A structured state-space model updates an explicit hidden state through causal state evolution:
\begin{equation}
\label{eq:ssm_form}
h_t = A_t h_{t-1} + B_t z_t,
\qquad
\hat{x}_t = C_t h_t,
\end{equation}
where $z_t$ denotes the observation-control input. This form closely matches Equation~\eqref{eq:belief_recursion}: it propagates latent memory and injects new information at each step.

A recurrent neural network also has a recursive form,
\begin{equation}
\label{eq:rnn_form}
h_t = \sigma(W_h h_{t-1} + W_z z_t + b),
\qquad
\hat{x}_t = W_o h_t.
\end{equation}
RNNs are therefore structurally closer to filtering than non-recursive architectures, but they entangle prior propagation and observation correction inside a generic nonlinear map. This gives them a weaker bias toward separating latent dynamics from observation injection.

Transformers aggregate the causal prefix through attention,
\begin{equation}
\label{eq:transformer_form}
o_t =
W_V
\sum_{s\le t}\alpha_{s,t}\xi_s,
\qquad
\alpha_{s,t}
=
\frac{\exp(q_t^\top k_s/\sqrt{d_k})}
{\sum_{\tau\le t}\exp(q_t^\top k_\tau/\sqrt{d_k})}.
\end{equation}
This computation is expressive and can represent filtering-like behavior, but it does not explicitly impose recursive latent-state compression. Under finite data and fixed parameter budgets, the model must learn both what information to use and how to compress and propagate it stably.

\subsection{Implications for Learned Nonlinear Filtering}

This analysis is an inductive-bias interpretation, not a formal separation result. Transformers and RNNs can implement filtering-like computations, and attention-based models can represent Kalman filtering in linear settings. Our claim is instead that selective SSMs make recursive latent-state propagation a default computation, which can make filtering behavior easier to learn under finite data, fixed parameter budgets, and long horizons.

This suggests three reasons why SSMs perform well in our benchmark. First, filtering is a recursive latent-state estimation problem, and SSMs directly parameterize this operator class. Second, SSM updates naturally combine latent propagation with observation injection. Third, long-horizon evaluation requires stable repeated composition, which is well matched to explicit state evolution.

Thus, SSMs provide a useful inductive bias for learned nonlinear state estimation, but this does not imply that Transformers or RNNs cannot learn filters. The empirical results suggest that, in the finite-data and fixed-budget regime studied here, SSMs are more readily trained into accurate and stable causal estimators than the other neural architectures considered.

\section{Conclusions}
\label{sec:conclusions}

We presented a systematic empirical comparison between neural state estimators and classical nonlinear filters across five nonlinear dynamical systems. Under a unified protocol, we evaluated recurrent networks, Transformer-based models, structured state-space models, and classical filters.

Neural sequence models trained from supervised data can perform effective causal state estimation without access to system dynamics, observation functions, or noise parameters. Among the model-free neural estimators studied here, structured state-space models achieve the strongest overall accuracy under the fixed parameter budget. They are competitive with strong classical filters in several nonlinear systems, but accurate model-based filters can still dominate when their assumptions are well matched to the system.

Our inductive-bias analysis links these results to computation structure. Filtering is fundamentally recursive latent-state estimation, and SSMs directly parameterize causal state evolution with observation injection. This structure plausibly explains their strong finite-data and long-horizon performance relative to generic recurrence and causal attention in our experiments. At the same time, our results should be interpreted as evidence about supervised point-estimation performance, not as a complete replacement for Bayesian filtering with calibrated uncertainty.

\textbf{Limitations.}
Our benchmark relies on clean simulator state labels for supervised training, so it does not address observation-only regimes where latent states are unavailable. Extending these comparisons to self-supervised, likelihood-based, or latent-variable training is an important direction.

The evaluation focuses on point-estimation accuracy, stability, robustness, and runtime. It does not study uncertainty calibration, posterior quality, or downstream closed-loop control performance.

Finally, our comparison focuses on fully model-free sequence estimators and classical model-based filters. Hybrid learned filters, such as neural Kalman-style estimators or differentiable particle-filter variants, occupy an important intermediate regime and are complementary to the present study.

\bibliography{references}
\bibliographystyle{plainnat}


\appendix

\section{Hyperparameter Details}
\label{app:hparams}

All neural models use the same optimization setup. Training uses AdamW with learning rate $2\times10^{-3}$, weight decay $10^{-5}$, and gradient clipping with maximum norm $1.0$. A ReduceLROnPlateau scheduler with decay factor $0.5$ and patience $3$ adapts the learning rate during training.

Models are trained for up to $500$ epochs with batch size $64$. Early stopping with patience $30$ is applied based on validation performance. The first $10$ timesteps are ignored during loss computation as a warmup period, and validation is performed after each epoch.

Table~\ref{tab:hparams} summarizes the hyperparameters used for the classical filters and neural models.

\begin{table}[htbp]
  \centering
  \caption{Hyperparameters for classical filters and neural models.}
  \label{tab:hparams}
  \renewcommand{\arraystretch}{1.08}
  \begin{tabular}{p{0.18\textwidth}p{0.76\textwidth}}
    \toprule
    Method & Hyperparameters \\
    \midrule
    \multicolumn{2}{l}{\textit{Classical Filters}} \\
    \midrule
    EKF & standard configuration \\
    UKF & $\alpha=10^{-3}$, $\beta=2$, $\kappa=0$ \\
    EnKF & ensemble size $=50$ \\
    PF & particles $=1024$, $x_0$ std $=1.0$, ESS fraction $=0.5$, rejuvenation std $=0.0$ \\
    \midrule
    \multicolumn{2}{l}{\textit{Neural Models}} \\
    \midrule
    GRU & hidden size $=85$, layers $=2$, dropout $=0.1$ \\
    LSTM & hidden size $=75$, layers $=2$, dropout $=0.1$ \\
    GPT-2 & layers $=2$, heads $=4$, embedding dim $=60$, positions $=128$, dropout $=0.1$, decode chunk length $=256$ \\
    Filterformer & encoding dim $=108$, heads $=4$, dropout $=0.1$ \\
    Mamba & $d_{\mathrm{model}}=80$, $d_{\mathrm{state}}=16$, $d_{\mathrm{conv}}=4$, expand $=2$, layers $=2$ \\
    Mamba-2 & $d_{\mathrm{model}}=64$, $d_{\mathrm{state}}=48$, $d_{\mathrm{conv}}=4$, expand $=2$, head dim $=16$, layers $=3$ \\
    \bottomrule
  \end{tabular}
\end{table}

\section{Full Evaluation Metrics}
\label{app:full_metrics}

Tables~\ref{tab:ballistic_reentry_extended}--\ref{tab:planar_quadrotor_extended} report detailed metrics for each method and scenario. Each entry is reported as median $\pm$ interquartile range unless otherwise indicated. The column ``Full RMSE'' denotes RMSE computed before trajectory-level outlier removal, while ``RMSE'' denotes the main aggregate RMSE after applying the outlier rule described in Section~\ref{sec:methodology}.

\begin{table}[htbp]
  \centering
  \caption{Extended metrics for Ballistic Re-entry.}
  \label{tab:ballistic_reentry_extended}
  \footnotesize
  \setlength{\tabcolsep}{1.6pt}
  \renewcommand{\arraystretch}{1.05}
  \begin{tabular*}{\textwidth}{@{\extracolsep{\fill}}lcccccccc}
    \toprule
    Method & RMSE & Full RMSE & NRMSE & MAE & MedAE & AUC & Out. & Drift \\
    \midrule
    \multicolumn{9}{l}{\textit{Classical Filters}} \\
    EKF & 5.289$\pm$0.004 & 5.289$\pm$0.004 & 1.571$\pm$0.005 & 3.008$\pm$0.003 & 0.900$\pm$0.003 & 2367.7$\pm$1.1 & \textbf{0.000} & 3.763 \\
    UKF & \textbf{0.626$\pm$0.010} & \textbf{0.628$\pm$0.009} & \textbf{0.607$\pm$0.004} & \textbf{0.361$\pm$0.003} & 0.154$\pm$0.000 & \textbf{303.9$\pm$4.9} & \textbf{0.000} & \textbf{1.217} \\
    EnKF & 0.821$\pm$0.062 & 2.166$\pm$0.068 & 0.878$\pm$0.009 & 0.560$\pm$0.004 & 0.159$\pm$0.000 & 391.7$\pm$28.2 & 0.022 & 2.708 \\
    PF & 0.657$\pm$0.010 & 0.779$\pm$0.037 & 0.648$\pm$0.005 & 0.366$\pm$0.004 & \textbf{0.145$\pm$0.000} & 317.4$\pm$5.1 & 0.002 & 1.762 \\
    \midrule
    \multicolumn{9}{l}{\textit{Neural Models}} \\
    GRU & 0.676$\pm$0.058 & 0.676$\pm$0.057 & 0.629$\pm$0.054 & 0.425$\pm$0.041 & 0.243$\pm$0.024 & 324.9$\pm$24.0 & \textbf{0.000} & 1.511 \\
    LSTM & 0.715$\pm$0.036 & 0.715$\pm$0.036 & 0.699$\pm$0.038 & 0.427$\pm$0.031 & 0.243$\pm$0.041 & 339.5$\pm$16.0 & \textbf{0.000} & 1.624 \\
    GPT-2 & 0.759$\pm$0.028 & 0.759$\pm$0.024 & 0.696$\pm$0.028 & 0.469$\pm$0.019 & 0.280$\pm$0.026 & 360.6$\pm$14.0 & \textbf{0.000} & 1.616 \\
    Filterformer & 0.793$\pm$0.098 & 0.804$\pm$0.090 & 0.717$\pm$0.059 & 0.532$\pm$0.062 & 0.333$\pm$0.050 & 383.3$\pm$43.6 & 0.001 & 1.471 \\
    Mamba & 0.654$\pm$0.039 & 0.663$\pm$0.037 & 0.613$\pm$0.018 & 0.423$\pm$0.018 & 0.243$\pm$0.021 & 315.9$\pm$17.5 & 0.001 & \textbf{1.382} \\
    Mamba-2 & \textbf{0.631$\pm$0.049} & \textbf{0.631$\pm$0.049} & \textbf{0.593$\pm$0.029} & \textbf{0.384$\pm$0.049} & \textbf{0.203$\pm$0.044} & \textbf{306.2$\pm$22.0} & \textbf{0.000} & 1.425 \\
    \bottomrule
  \end{tabular*}
\end{table}

\begin{table}[htbp]
  \centering
  \caption{Extended metrics for Bearings-only Tracking. Entries reported as $\infty$ denote numerical divergence of EnKF.}
  \label{tab:bearings_only_tracking_extended}
  \footnotesize
  \setlength{\tabcolsep}{0.8pt}
  \renewcommand{\arraystretch}{1.05}
  \begin{tabular*}{\textwidth}{@{\extracolsep{\fill}}lcccccccc}
    \toprule
    Method & RMSE & Full RMSE & NRMSE & MAE & MedAE & AUC & Out. & Drift \\
    \midrule
    \multicolumn{9}{l}{\textit{Classical Filters}} \\
    EKF & 308.624$\pm$2.991 & 308.624$\pm$2.991 & 1.005$\pm$0.005 & 146.692$\pm$1.311 & 4.441$\pm$0.024 & 127066.4$\pm$1318.8 & \textbf{0.000} & 8.600 \\
    UKF & \textbf{180.946$\pm$2.524} & \textbf{180.946$\pm$2.524} & \textbf{0.647$\pm$0.005} & \textbf{73.398$\pm$0.881} & \textbf{2.751$\pm$0.026} & \textbf{76324.1$\pm$1269.0} & \textbf{0.000} & 6.877 \\
    EnKF & $\infty$ & $\infty$ & $\infty$ & $\infty$ & $\infty$ & $\infty$ & \textit{N/A} & \textit{N/A} \\
    PF & 362.554$\pm$1.487 & 362.554$\pm$1.487 & 1.429$\pm$0.013 & 108.652$\pm$0.413 & 4.190$\pm$0.013 & 137847.6$\pm$812.5 & \textbf{0.000} & 17.660 \\
    \midrule
    \multicolumn{9}{l}{\textit{Neural Models}} \\
    GRU & 191.896$\pm$4.097 & 191.896$\pm$4.097 & 0.637$\pm$0.016 & 92.734$\pm$2.817 & 3.225$\pm$0.115 & 86488.9$\pm$2641.6 & \textbf{0.000} & 2.473 \\
    LSTM & 187.855$\pm$12.451 & 187.855$\pm$12.451 & 0.616$\pm$0.034 & 88.824$\pm$12.034 & 3.144$\pm$0.430 & 82921.0$\pm$7766.4 & \textbf{0.000} & 2.929 \\
    GPT-2 & 225.685$\pm$4.852 & 225.685$\pm$4.852 & 0.699$\pm$0.009 & 103.365$\pm$1.901 & 3.310$\pm$0.133 & 95102.8$\pm$2103.0 & \textbf{0.000} & 3.396 \\
    Filterformer & 224.502$\pm$4.744 & 224.502$\pm$4.744 & 0.744$\pm$0.009 & 115.997$\pm$4.808 & 3.792$\pm$0.121 & 101111.5$\pm$2203.7 & \textbf{0.000} & \textbf{2.048} \\
    Mamba & \textbf{168.081$\pm$9.883} & \textbf{168.081$\pm$9.883} & \textbf{0.588$\pm$0.021} & \textbf{82.448$\pm$7.004} & \textbf{2.981$\pm$0.128} & \textbf{78788.7$\pm$5939.4} & \textbf{0.000} & 2.138 \\
    Mamba-2 & 177.842$\pm$26.087 & 177.842$\pm$26.087 & 0.611$\pm$0.065 & 85.200$\pm$15.568 & 3.049$\pm$0.323 & 81010.2$\pm$12755.1 & \textbf{0.000} & 2.474 \\
    \bottomrule
  \end{tabular*}
\end{table}

\begin{table}[htbp]
  \centering
  \caption{Extended metrics for Lorenz 96.}
  \label{tab:l96_extended}
  \footnotesize
  \setlength{\tabcolsep}{1.6pt}
  \renewcommand{\arraystretch}{1.05}
  \begin{tabular*}{\textwidth}{@{\extracolsep{\fill}}lcccccccc}
    \toprule
    Method & RMSE & Full RMSE & NRMSE & MAE & MedAE & AUC & Out. & Drift \\
    \midrule
    \multicolumn{9}{l}{\textit{Classical Filters}} \\
    EKF & 0.049$\pm$0.000 & 0.049$\pm$0.000 & 0.014$\pm$0.000 & 0.027$\pm$0.000 & \textbf{0.013$\pm$0.000} & 23.7$\pm$0.0 & \textbf{0.000} & 1.126 \\
    UKF & \textbf{0.048$\pm$0.000} & \textbf{0.048$\pm$0.000} & \textbf{0.013$\pm$0.000} & \textbf{0.026$\pm$0.000} & \textbf{0.013$\pm$0.000} & \textbf{23.2$\pm$0.0} & \textbf{0.000} & 1.125 \\
    EnKF & 0.054$\pm$0.000 & 0.054$\pm$0.000 & 0.015$\pm$0.000 & 0.029$\pm$0.000 & \textbf{0.013$\pm$0.000} & 26.1$\pm$0.0 & \textbf{0.000} & 1.125 \\
    PF & 0.065$\pm$0.000 & 0.065$\pm$0.000 & 0.018$\pm$0.000 & 0.037$\pm$0.000 & 0.018$\pm$0.000 & 31.6$\pm$0.1 & \textbf{0.000} & \textbf{1.087} \\
    \midrule
    \multicolumn{9}{l}{\textit{Neural Models}} \\
    GRU & 0.127$\pm$0.010 & 0.127$\pm$0.011 & 0.036$\pm$0.003 & 0.101$\pm$0.009 & 0.085$\pm$0.008 & 61.9$\pm$5.1 & 0.001 & \textbf{0.976} \\
    LSTM & 0.119$\pm$0.005 & 0.119$\pm$0.005 & 0.034$\pm$0.002 & 0.093$\pm$0.004 & 0.077$\pm$0.003 & 57.9$\pm$2.6 & 0.001 & 0.979 \\
    GPT-2 & 0.099$\pm$0.015 & 0.099$\pm$0.015 & 0.028$\pm$0.004 & 0.080$\pm$0.015 & 0.070$\pm$0.017 & 48.2$\pm$7.3 & \textbf{0.000} & 1.045 \\
    Filterformer & 0.732$\pm$0.061 & 0.732$\pm$0.061 & 0.202$\pm$0.018 & 0.493$\pm$0.038 & 0.327$\pm$0.029 & 340.9$\pm$24.8 & \textbf{0.000} & 2.365 \\
    Mamba & \textbf{0.052$\pm$0.001} & \textbf{0.052$\pm$0.001} & \textbf{0.015$\pm$0.000} & \textbf{0.031$\pm$0.001} & \textbf{0.017$\pm$0.001} & 25.4$\pm$0.5 & \textbf{0.000} & 1.092 \\
    Mamba-2 & \textbf{0.052$\pm$0.001} & \textbf{0.052$\pm$0.001} & \textbf{0.015$\pm$0.000} & \textbf{0.031$\pm$0.001} & \textbf{0.017$\pm$0.001} & \textbf{25.4$\pm$0.3} & \textbf{0.000} & 1.067 \\
    \bottomrule
  \end{tabular*}
\end{table}

\begin{table}[htbp]
  \centering
  \caption{Extended metrics for N-link Pendulum.}
  \label{tab:nlink_pendulum_extended}
  \footnotesize
  \setlength{\tabcolsep}{1.6pt}
  \renewcommand{\arraystretch}{1.05}
  \begin{tabular*}{\textwidth}{@{\extracolsep{\fill}}lcccccccc}
    \toprule
    Method & RMSE & Full RMSE & NRMSE & MAE & MedAE & AUC & Out. & Drift \\
    \midrule
    \multicolumn{9}{l}{\textit{Classical Filters}} \\
    EKF & \textbf{0.080$\pm$0.000} & \textbf{0.081$\pm$0.000} & \textbf{0.012$\pm$0.000} & \textbf{0.063$\pm$0.000} & \textbf{0.053$\pm$0.000} & \textbf{38.9$\pm$0.0} & 0.005 & 1.063 \\
    UKF & 0.081$\pm$0.000 & \textbf{0.081$\pm$0.000} & \textbf{0.012$\pm$0.000} & 0.065$\pm$0.000 & 0.054$\pm$0.000 & 39.6$\pm$0.0 & 0.001 & 1.042 \\
    EnKF & 0.092$\pm$0.000 & 0.092$\pm$0.000 & 0.014$\pm$0.000 & 0.073$\pm$0.000 & 0.062$\pm$0.000 & 45.2$\pm$0.0 & \textbf{0.000} & \textbf{1.012} \\
    PF & 0.086$\pm$0.000 & 4.678$\pm$0.034 & 0.550$\pm$0.005 & 0.618$\pm$0.002 & 0.060$\pm$0.000 & 42.2$\pm$0.0 & 0.272 & 71.143 \\
    \midrule
    \multicolumn{9}{l}{\textit{Neural Models}} \\
    GRU & 0.399$\pm$0.051 & 0.452$\pm$0.044 & 0.063$\pm$0.006 & 0.242$\pm$0.030 & 0.159$\pm$0.016 & 175.1$\pm$20.4 & 0.004 & 4.039 \\
    LSTM & 0.356$\pm$0.011 & 0.421$\pm$0.020 & 0.058$\pm$0.003 & 0.217$\pm$0.004 & 0.142$\pm$0.004 & 157.9$\pm$4.1 & 0.003 & 4.101 \\
    GPT-2 & 0.233$\pm$0.043 & 0.269$\pm$0.027 & 0.037$\pm$0.005 & 0.153$\pm$0.017 & 0.106$\pm$0.004 & 104.2$\pm$17.1 & 0.004 & 3.474 \\
    Filterformer & 1.749$\pm$0.027 & 1.760$\pm$0.037 & 0.219$\pm$0.005 & 0.701$\pm$0.014 & 0.252$\pm$0.025 & 770.5$\pm$12.5 & \textbf{0.000} & 3.655 \\
    Mamba & 0.093$\pm$0.003 & 0.118$\pm$0.013 & 0.017$\pm$0.002 & 0.076$\pm$0.004 & 0.062$\pm$0.002 & 45.6$\pm$1.4 & 0.054 & 1.847 \\
    Mamba-2 & \textbf{0.092$\pm$0.001} & \textbf{0.105$\pm$0.010} & \textbf{0.015$\pm$0.001} & \textbf{0.073$\pm$0.001} & \textbf{0.061$\pm$0.000} & \textbf{44.8$\pm$0.2} & 0.020 & \textbf{1.366} \\
    \bottomrule
  \end{tabular*}
\end{table}

\begin{table}[htbp]
  \centering
  \caption{Extended metrics for Planar Quadrotor.}
  \label{tab:planar_quadrotor_extended}
  \footnotesize
  \setlength{\tabcolsep}{1.6pt}
  \renewcommand{\arraystretch}{1.05}
  \begin{tabular*}{\textwidth}{@{\extracolsep{\fill}}lcccccccc}
    \toprule
    Method & RMSE & Full RMSE & NRMSE & MAE & MedAE & AUC & Out. & Drift \\
    \midrule
    \multicolumn{9}{l}{\textit{Classical Filters}} \\
    EKF & \textbf{0.287$\pm$0.005} & \textbf{0.288$\pm$0.005} & \textbf{0.009$\pm$0.000} & \textbf{0.132$\pm$0.001} & \textbf{0.044$\pm$0.000} & \textbf{53.7$\pm$0.9} & \textbf{0.000} & \textbf{1.415} \\
    UKF & 8.042$\pm$0.125 & 8.042$\pm$0.125 & 0.073$\pm$0.001 & 1.000$\pm$0.005 & 0.057$\pm$0.000 & 1213.6$\pm$9.0 & \textbf{0.000} & 15.425 \\
    EnKF & 9.960$\pm$0.156 & 9.960$\pm$0.156 & 0.092$\pm$0.001 & 1.795$\pm$0.017 & 0.099$\pm$0.000 & 1540.3$\pm$21.2 & \textbf{0.000} & 14.995 \\
    PF & 128.009$\pm$3.969 & 114.478$\pm$2.873 & 0.957$\pm$0.005 & 38.001$\pm$0.325 & 1.453$\pm$0.039 & 18038.7$\pm$409.9 & 0.187 & 14.263 \\
    \midrule
    \multicolumn{9}{l}{\textit{Neural Models}} \\
    GRU & 13.023$\pm$1.843 & 36.532$\pm$4.522 & 0.224$\pm$0.041 & 6.170$\pm$0.633 & 1.437$\pm$0.148 & 2202.3$\pm$322.0 & 0.004 & 8.349 \\
    LSTM & 11.444$\pm$1.941 & 35.106$\pm$5.665 & 0.219$\pm$0.044 & 5.588$\pm$0.784 & 1.301$\pm$0.239 & 2013.4$\pm$322.2 & 0.006 & \textbf{7.716} \\
    GPT-2 & 16.420$\pm$2.358 & 55.501$\pm$2.790 & 0.262$\pm$0.026 & 8.209$\pm$1.086 & 1.856$\pm$0.335 & 2826.3$\pm$452.3 & 0.007 & 12.018 \\
    Filterformer & 14.345$\pm$1.270 & 40.923$\pm$6.972 & 0.254$\pm$0.045 & 6.285$\pm$0.626 & 1.595$\pm$0.117 & 2364.3$\pm$205.6 & \textbf{0.002} & 11.213 \\
    Mamba & 7.395$\pm$1.171 & 32.325$\pm$7.591 & 0.183$\pm$0.047 & 2.980$\pm$0.378 & 0.687$\pm$0.066 & 1310.0$\pm$238.1 & \textbf{0.002} & 9.787 \\
    Mamba-2 & \textbf{5.702$\pm$1.199} & \textbf{30.711$\pm$4.422} & \textbf{0.175$\pm$0.038} & \textbf{2.326$\pm$0.434} & \textbf{0.546$\pm$0.057} & \textbf{1015.1$\pm$203.6} & \textbf{0.002} & 10.077 \\
    \bottomrule
  \end{tabular*}
\end{table}

\section{Aggregate Accuracy and Stability Summaries}
\label{app:aggregate_summaries}

Table~\ref{tab:nrmse_avg_all} reports a compact cross-scenario summary of NRMSE. Each entry reports the aggregate NRMSE with the interquartile range in parentheses. This table complements the per-scenario results in Appendix~\ref{app:full_metrics}; because the benchmark systems differ in dynamics, observation structure, state dimension, and failure modes, the aggregate summary should be interpreted together with the detailed per-scenario tables.

\begin{table}[htbp]
  \centering
  \caption{Cross-scenario summary of NRMSE. Lower is better. The median and IQR are computed over the five scenario-level median NRMSE values for each method.}
  \label{tab:nrmse_avg_all}
  \resizebox{\textwidth}{!}{%
  \begin{tabular}{lcccc|cccccc}
    \toprule
    & \multicolumn{4}{c|}{\textit{Classical Filters}} & \multicolumn{6}{c}{\textit{Neural Models}} \\
    \cmidrule(lr){2-5} \cmidrule(lr){6-11}
    Metric & EKF & UKF & EnKF & PF & GRU & LSTM & GPT-2 & Filterformer & Mamba & Mamba-2 \\
    \midrule
    Median NRMSE & \textbf{0.014} & 0.073 & 0.092 & 0.648 & 0.224 & 0.219 & 0.262 & 0.254 & 0.183 & \textbf{0.175} \\
    (IQR) & (0.993) & (0.594) & (0.862) & (0.407) & (0.567) & (0.558) & (0.659) & (0.499) & (0.571) & (0.577) \\
    \bottomrule
  \end{tabular}}
\end{table}

Table~\ref{tab:stability_summary_app} reports per-scenario outlier and drift ratios. The outlier ratio measures the fraction of trajectories classified as outliers under the robust log-MSE rule, while the drift ratio measures error growth from the beginning to the end of the evaluation horizon. These metrics characterize long-horizon stability and catastrophic failures separately from point-estimation accuracy.

\begin{table}[htbp]
  \centering
  \caption{Per-scenario stability metrics. Out. denotes outlier ratio, and Drift denotes drift ratio. Lower is better. N/A indicates numerical divergence or unavailable statistics.}
  \label{tab:stability_summary_app}
  \footnotesize
  \setlength{\tabcolsep}{2.0pt}
  \renewcommand{\arraystretch}{1.05}
  \begin{tabular*}{\textwidth}{@{\extracolsep{\fill}}lccccc|ccccc}
    \toprule
    & \multicolumn{5}{c|}{Outlier Ratio}
    & \multicolumn{5}{c}{Drift Ratio} \\
    \cmidrule(lr){2-6}\cmidrule(lr){7-11}
    Method & Ball. & Bear. & L96 & N-Link & P. Quad.
           & Ball. & Bear. & L96 & N-Link & P. Quad. \\
    \midrule
    \multicolumn{11}{l}{\textit{Classical Filters}} \\
    EKF  & 0.000 & 0.000 & 0.000 & 0.005 & 0.000 & 3.763 & 8.600 & 1.126 & 1.063 & 1.415 \\
    UKF  & 0.000 & 0.000 & 0.000 & 0.001 & 0.000 & 1.217 & 6.877 & 1.125 & 1.042 & 15.425 \\
    EnKF & 0.022 & N/A   & 0.000 & 0.000 & 0.000 & 2.708 & N/A   & 1.125 & 1.012 & 14.995 \\
    PF   & 0.002 & 0.000 & 0.000 & 0.272 & 0.187 & 1.762 & 17.660 & 1.087 & 71.143 & 14.263 \\
    \midrule
    \multicolumn{11}{l}{\textit{Neural Models}} \\
    GRU          & 0.000 & 0.000 & 0.001 & 0.004 & 0.004 & 1.511 & 2.473 & 0.976 & 4.039 & 8.349 \\
    LSTM         & 0.000 & 0.000 & 0.001 & 0.003 & 0.006 & 1.624 & 2.929 & 0.979 & 4.101 & 7.716 \\
    GPT-2        & 0.000 & 0.000 & 0.000 & 0.004 & 0.007 & 1.616 & 3.396 & 1.045 & 3.474 & 12.018 \\
    Filterformer & 0.001 & 0.000 & 0.000 & 0.000 & 0.002 & 1.471 & 2.048 & 2.365 & 3.655 & 11.213 \\
    Mamba        & 0.001 & 0.000 & 0.000 & 0.054 & 0.002 & 1.382 & 2.138 & 1.092 & 1.847 & 9.787 \\
    Mamba-2      & 0.000 & 0.000 & 0.000 & 0.020 & 0.002 & 1.425 & 2.474 & 1.067 & 1.366 & 10.077 \\
    \bottomrule
  \end{tabular*}
\end{table}

\section{RMSE Over Time}
\label{app:rmse_over_time}

\begin{figure}[htbp]
  \centering
  \includegraphics[width=0.92\textwidth]{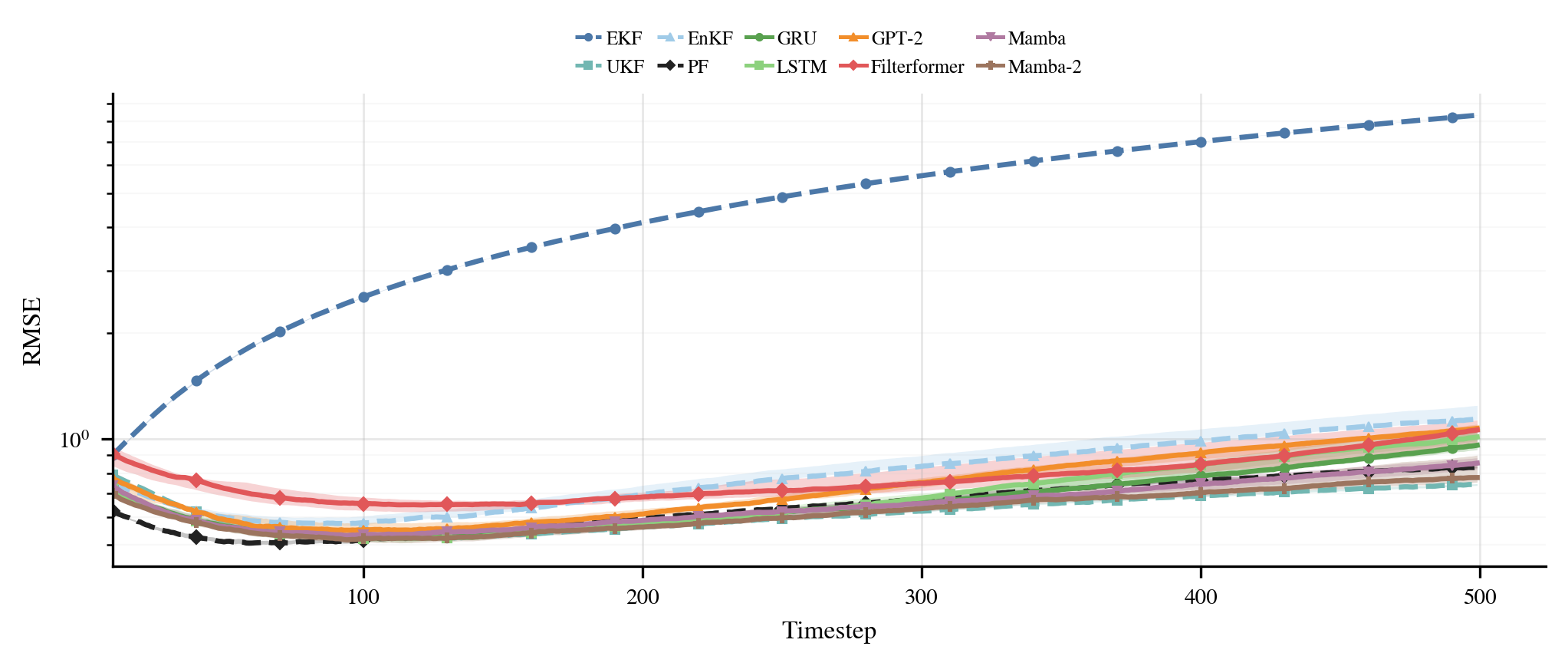}
  \caption{RMSE over time for Ballistic Re-entry. Curves show median RMSE with interquartile-range bands.}
  \label{fig:rmse_time_ballistic}
\end{figure}

\begin{figure}[htbp]
  \centering
  \includegraphics[width=0.92\textwidth]{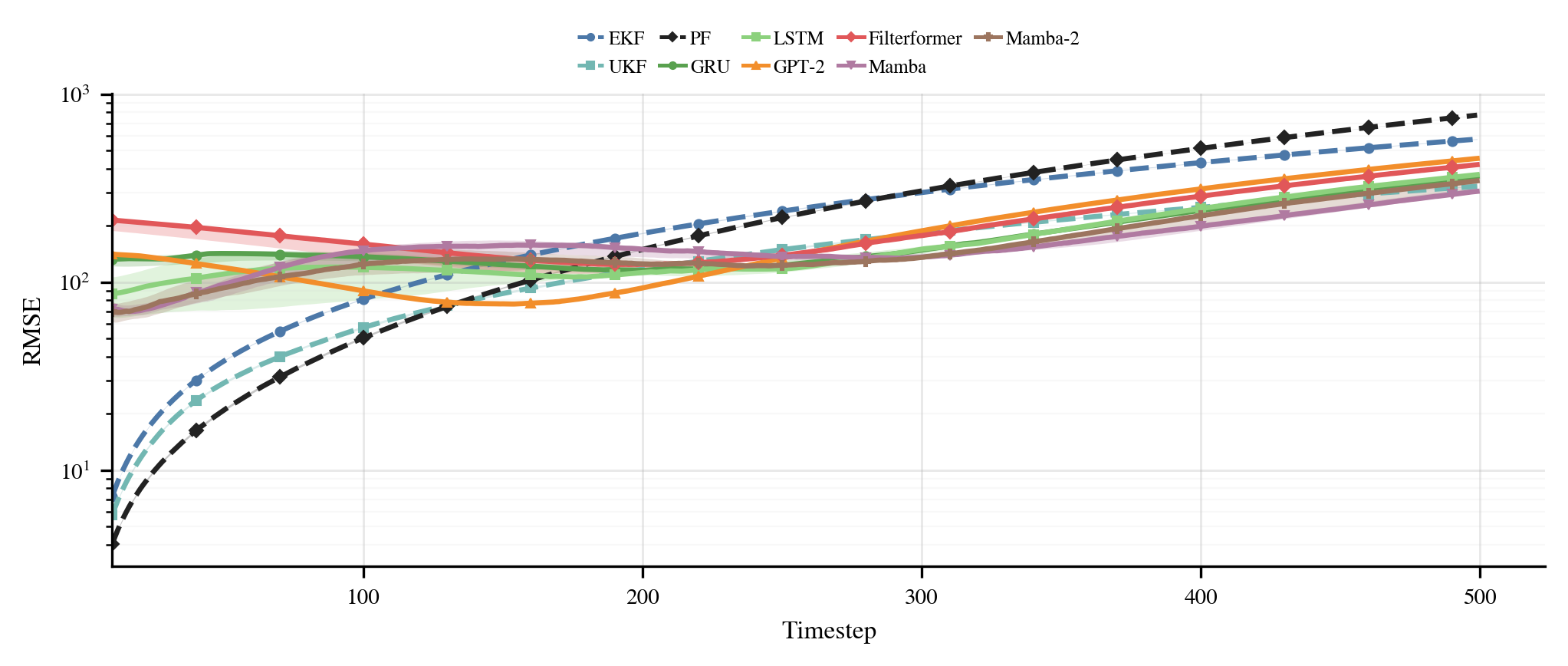}
  \caption{RMSE over time for Bearings-only Tracking. Curves show median RMSE with interquartile-range bands.}
  \label{fig:rmse_time_bearings}
\end{figure}

\begin{figure}[htbp]
  \centering
  \includegraphics[width=0.92\textwidth]{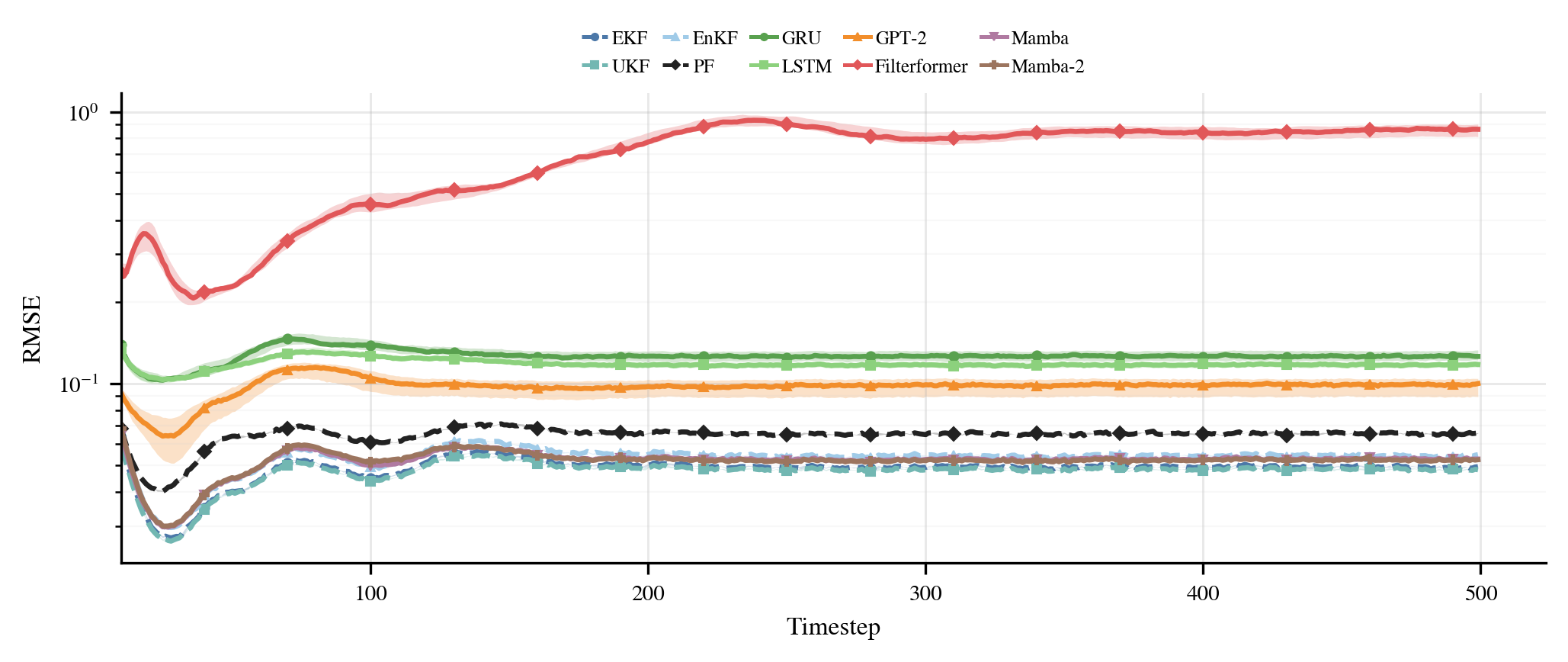}
  \caption{RMSE over time for Lorenz 96. Curves show median RMSE with interquartile-range bands.}
  \label{fig:rmse_time_l96}
\end{figure}

\begin{figure}[htbp]
  \centering
  \includegraphics[width=0.92\textwidth]{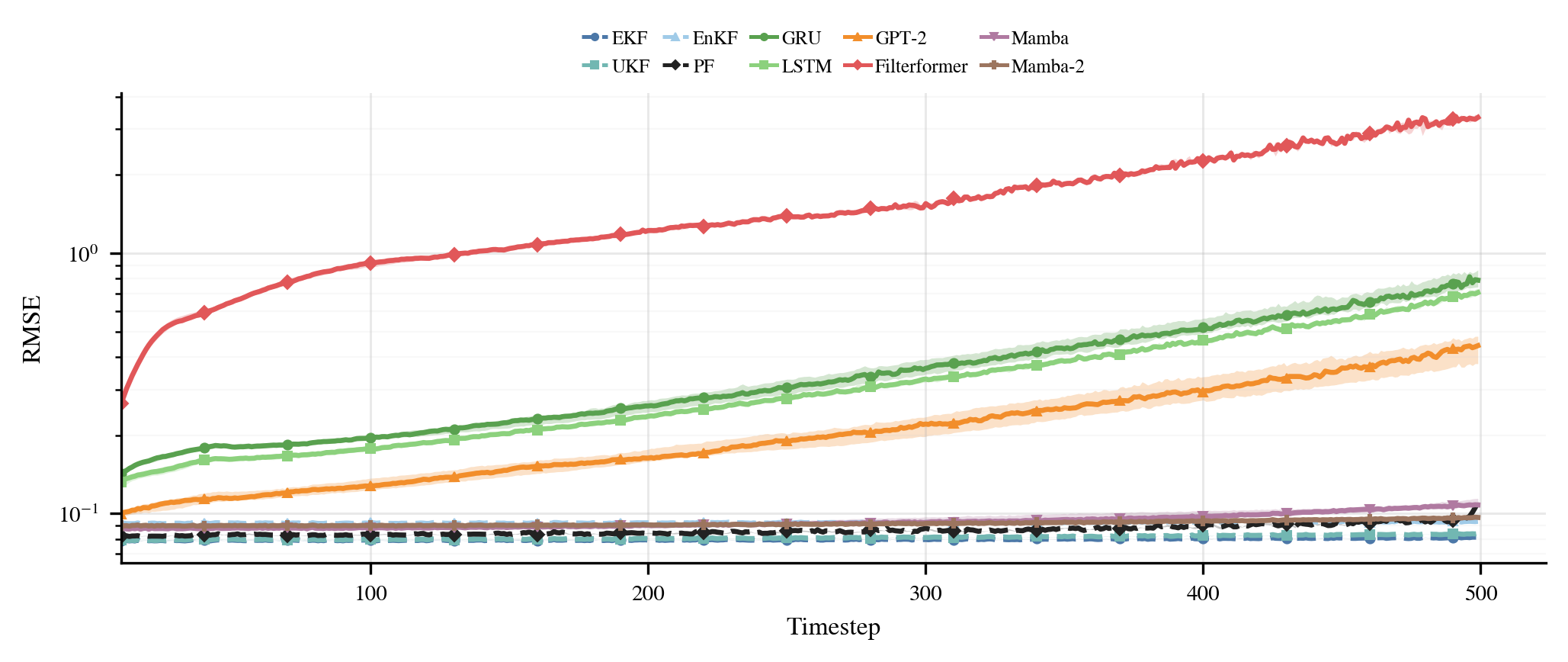}
  \caption{RMSE over time for N-link Pendulum. Curves show median RMSE with interquartile-range bands.}
  \label{fig:rmse_time_nlink}
\end{figure}

\begin{figure}[htbp]
  \centering
  \includegraphics[width=0.92\textwidth]{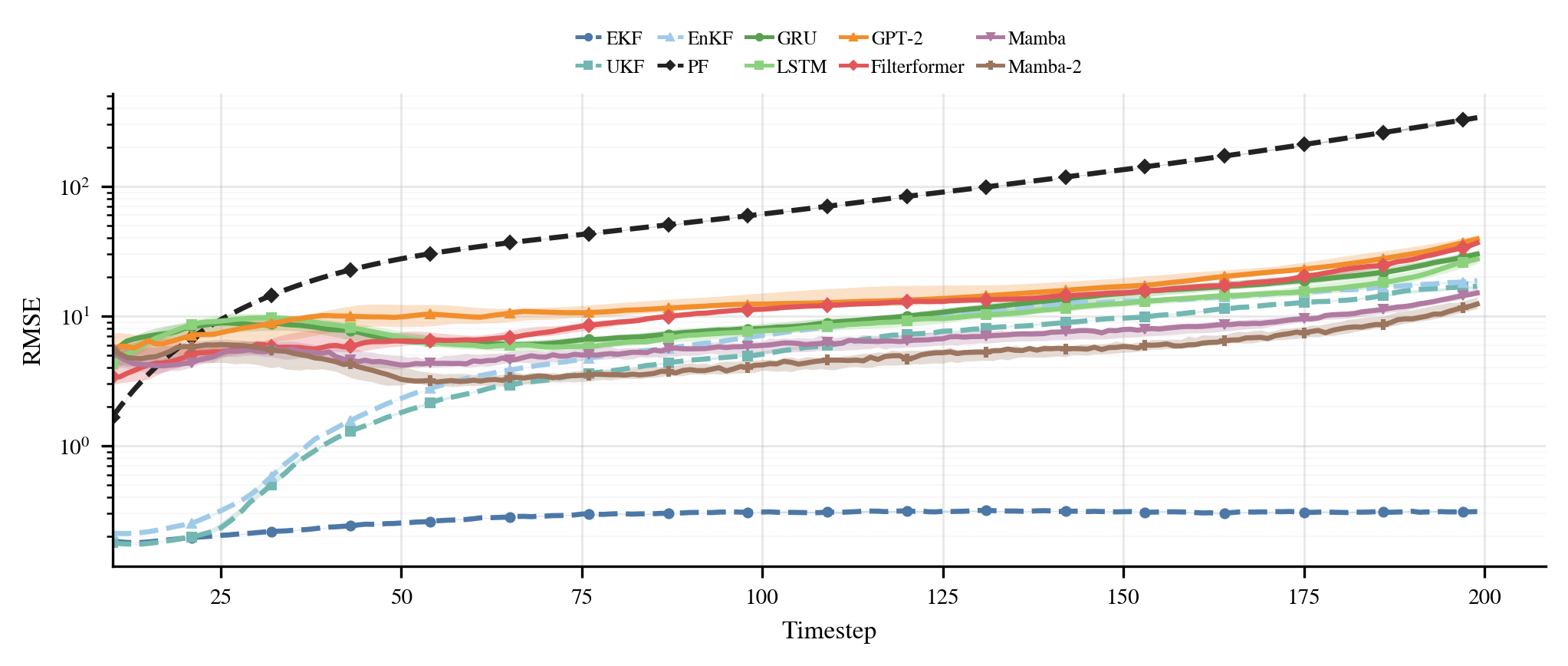}
  \caption{RMSE over time for Planar Quadrotor. Curves show median RMSE with interquartile-range bands.}
  \label{fig:rmse_time_quadrotor}
\end{figure}

Figures~\ref{fig:rmse_time_ballistic}--\ref{fig:rmse_time_quadrotor} report RMSE as a function of timestep for each scenario. Each curve corresponds to one method and shows the median RMSE across evaluation runs, with the interquartile range shown as an uncertainty band. These plots complement the aggregate metrics by showing how estimation errors evolve over the evaluation horizon.

\section{Robustness to Observation Noise}
\label{app:noise_robustness}

\begin{figure}[htbp]
  \centering
  \includegraphics[width=0.92\textwidth]{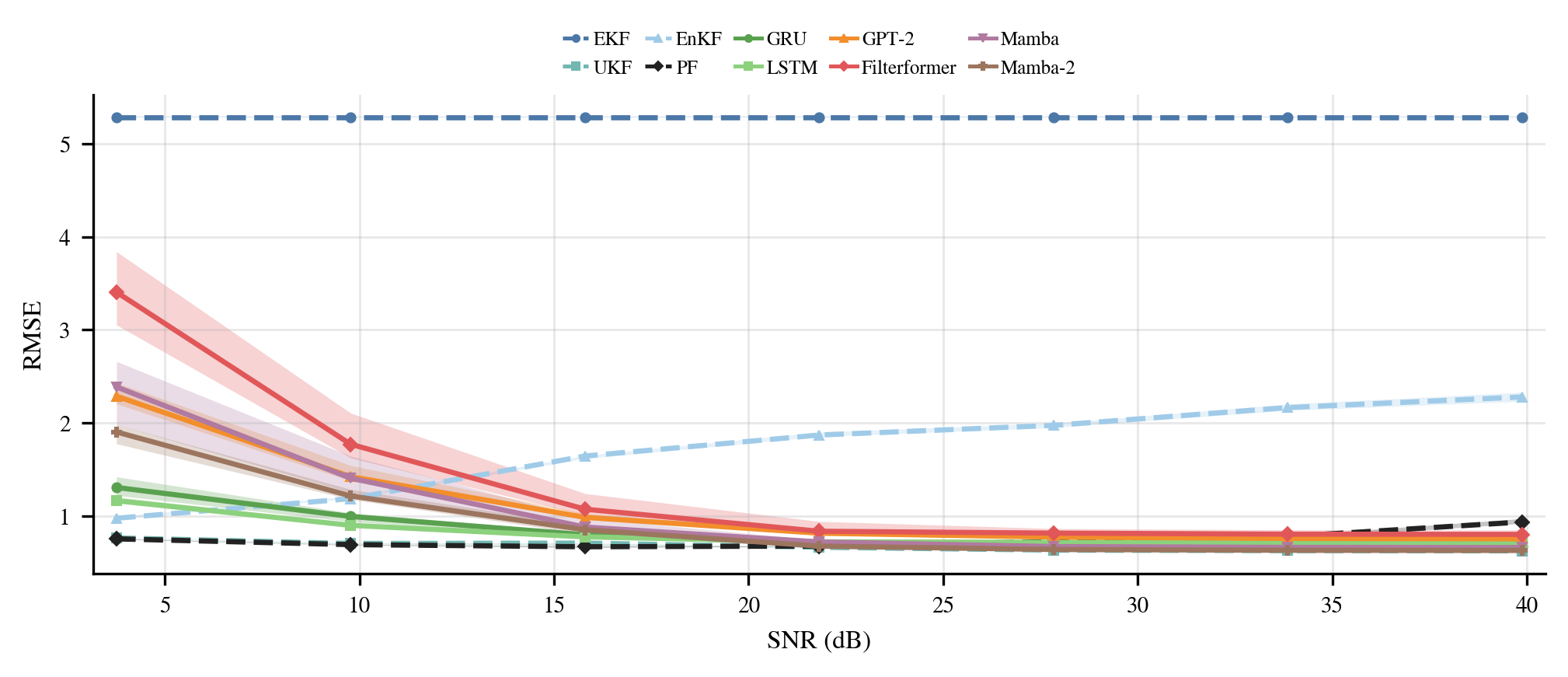}
  \caption{RMSE under different observation-noise levels for Ballistic Re-entry. Curves show median RMSE with interquartile-range bands.}
  \label{fig:noise_ballistic}
\end{figure}

\begin{figure}[htbp]
  \centering
  \includegraphics[width=0.92\textwidth]{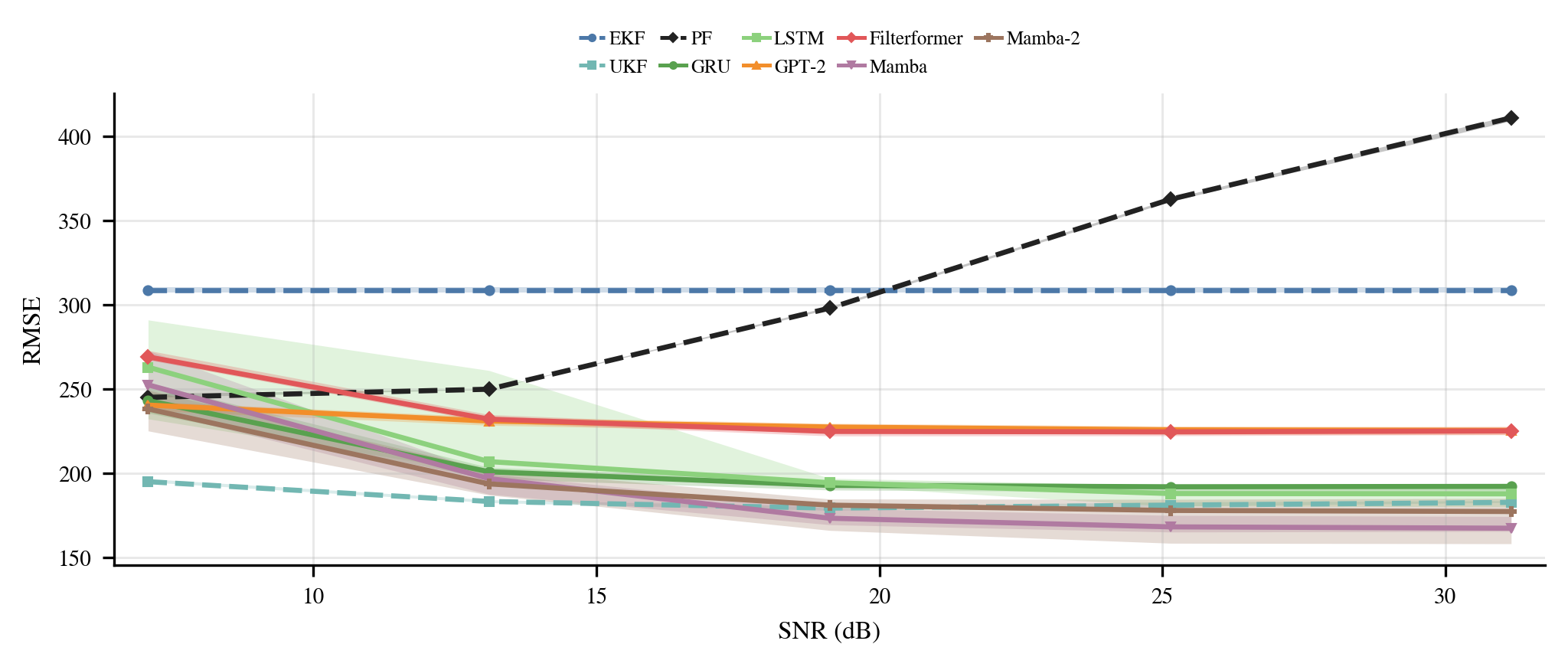}
  \caption{RMSE under different observation-noise levels for Bearings-only Tracking. Curves show median RMSE with interquartile-range bands.}
  \label{fig:noise_bearings}
\end{figure}

\begin{figure}[htbp]
  \centering
  \includegraphics[width=0.92\textwidth]{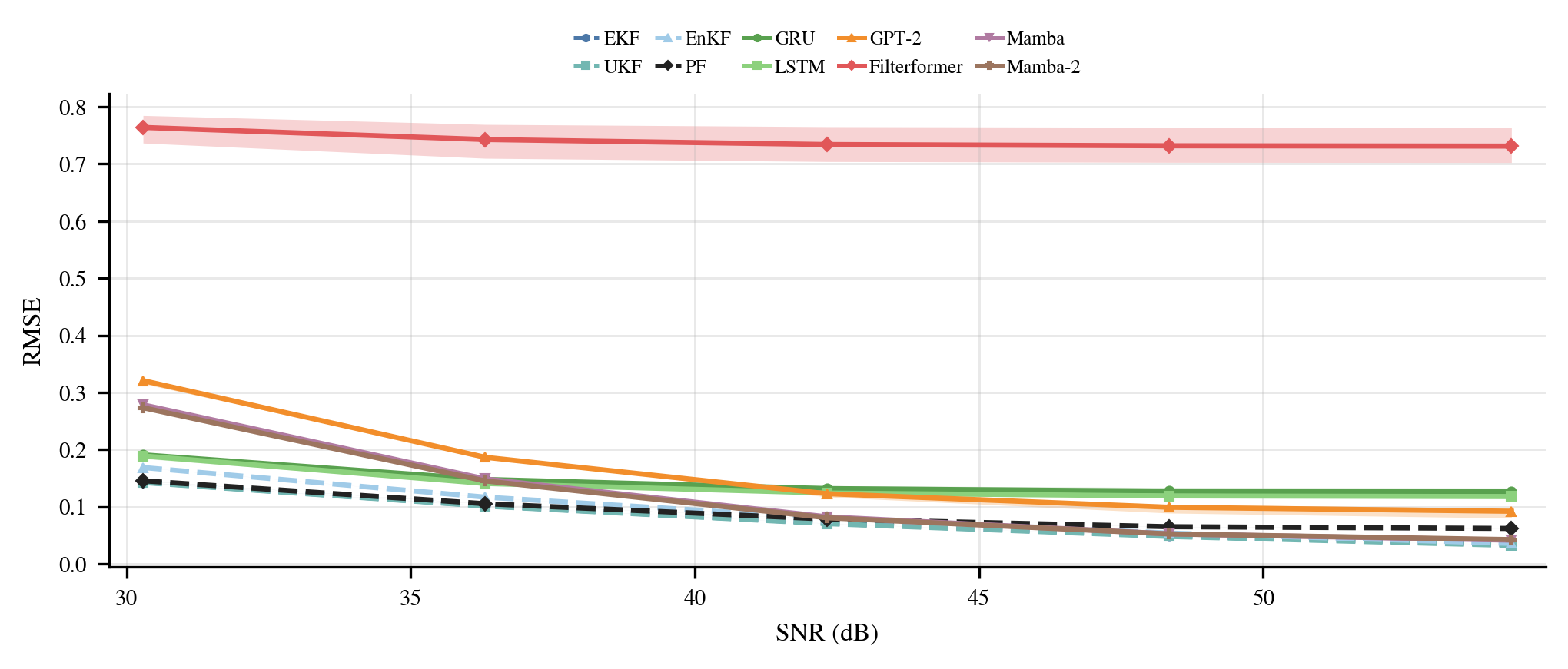}
  \caption{RMSE under different observation-noise levels for Lorenz 96. Curves show median RMSE with interquartile-range bands.}
  \label{fig:noise_l96}
\end{figure}

\begin{figure}[htbp]
  \centering
  \includegraphics[width=0.92\textwidth]{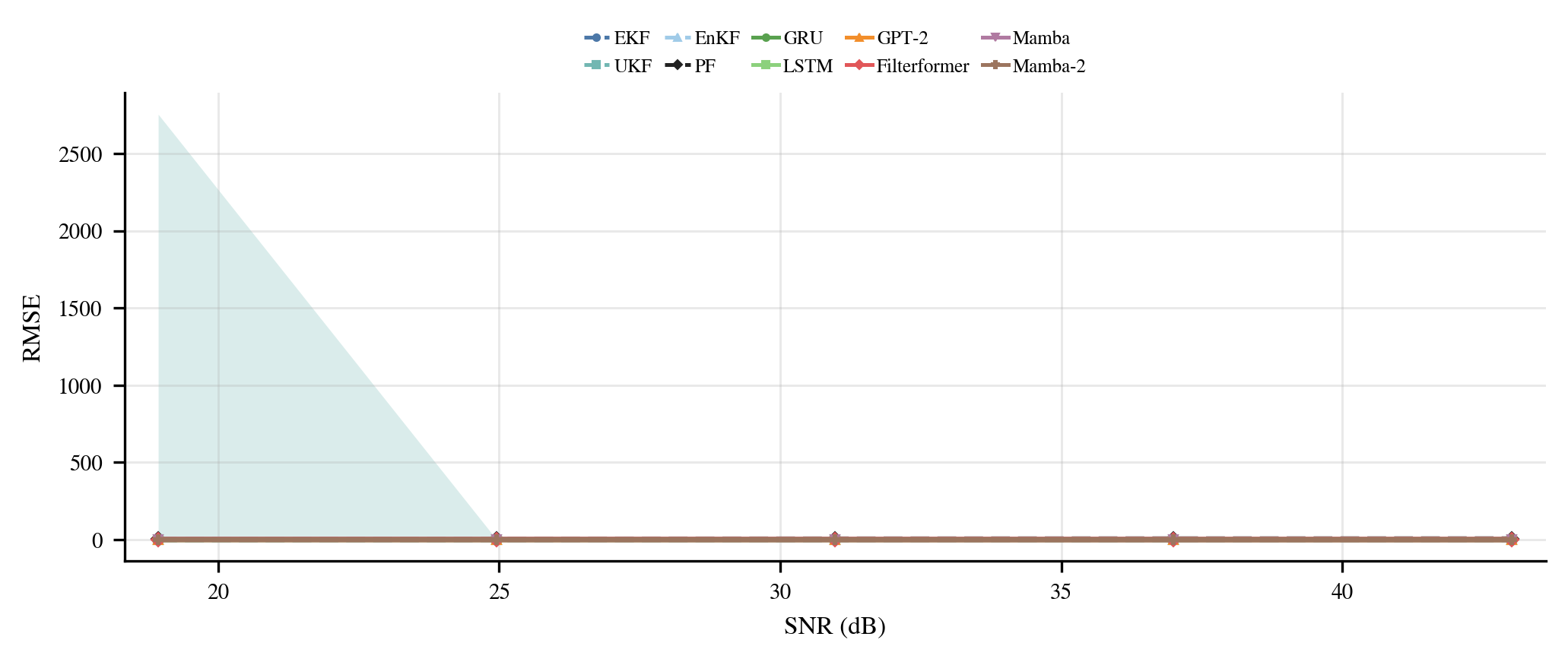}
  \caption{RMSE under different observation-noise levels for N-link Pendulum. Curves show median RMSE with interquartile-range bands.}
  \label{fig:noise_nlink}
\end{figure}

\begin{figure}[htbp]
  \centering
  \includegraphics[width=0.92\textwidth]{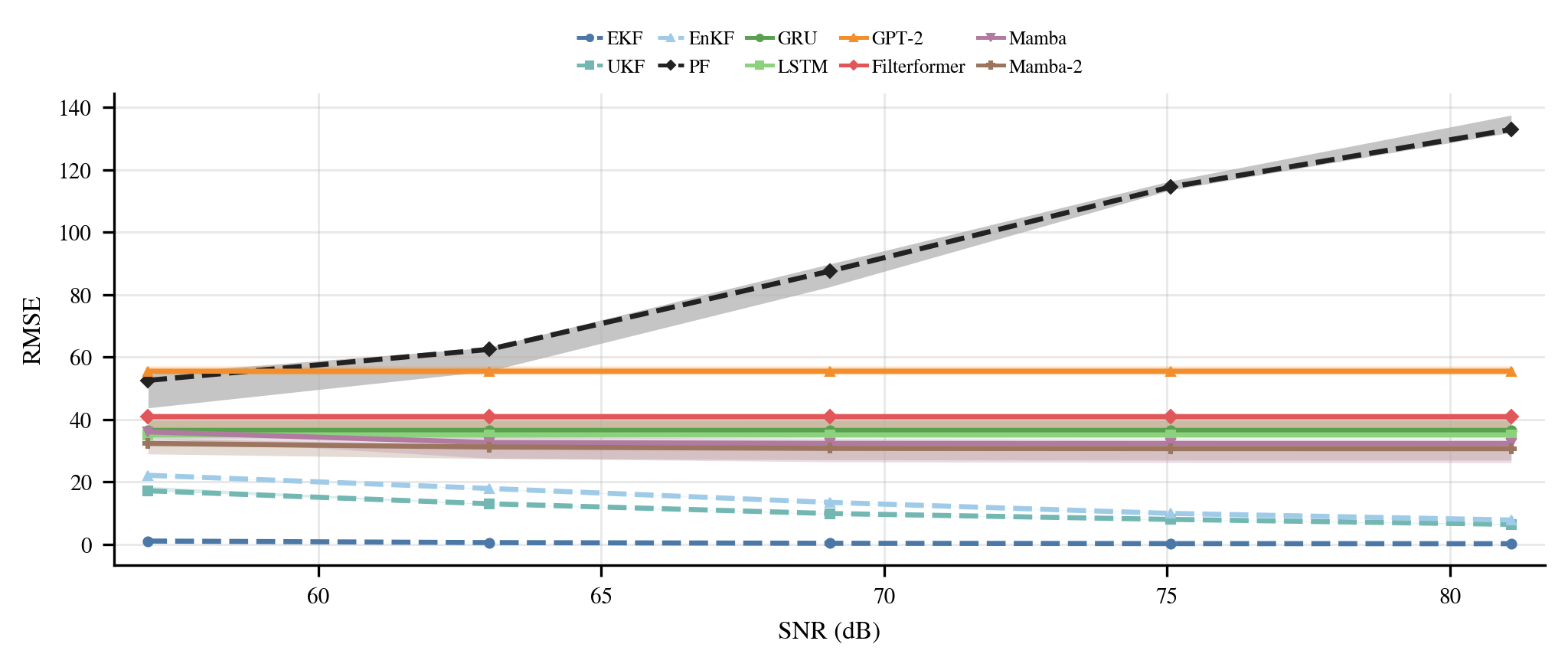}
  \caption{RMSE under different observation-noise levels for Planar Quadrotor. Curves show median RMSE with interquartile-range bands.}
  \label{fig:noise_quadrotor}
\end{figure}

Figures~\ref{fig:noise_ballistic}--\ref{fig:noise_quadrotor} report robustness under increased observation noise. For each scenario, RMSE is plotted as a function of the noise level or signal-to-noise ratio. Each curve corresponds to one method and shows the median RMSE across evaluation runs, with the interquartile range shown as an uncertainty band.

\section{Data Efficiency}
\label{app:data_efficiency}

\begin{figure}[htbp]
  \centering
  \includegraphics[width=0.92\textwidth]{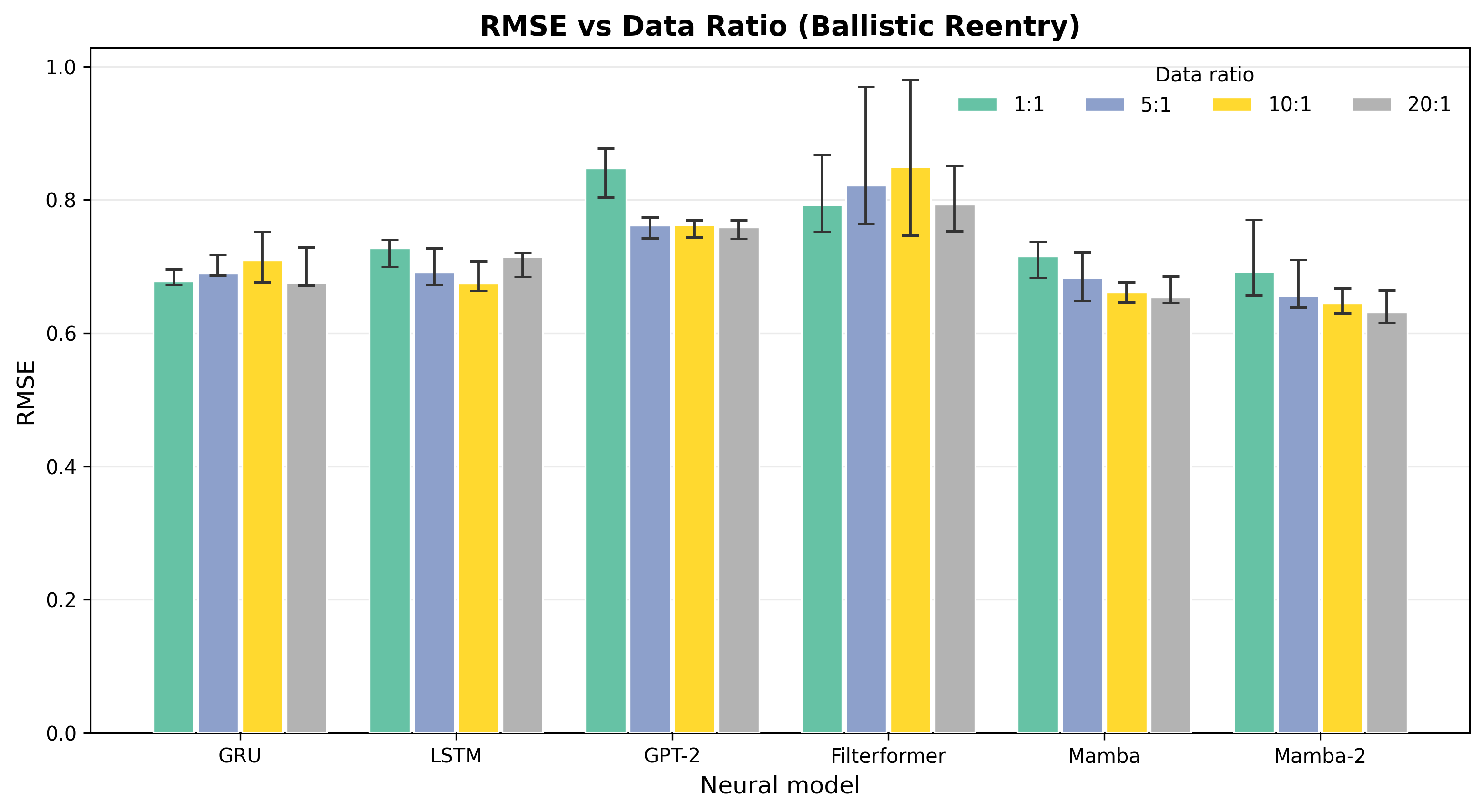}
  \caption{Effect of data-to-parameter ratio on Ballistic Re-entry. Bars show median performance with interquartile-range intervals.}
  \label{fig:data_ratio_ballistic}
\end{figure}

\begin{figure}[htbp]
  \centering
  \includegraphics[width=0.92\textwidth]{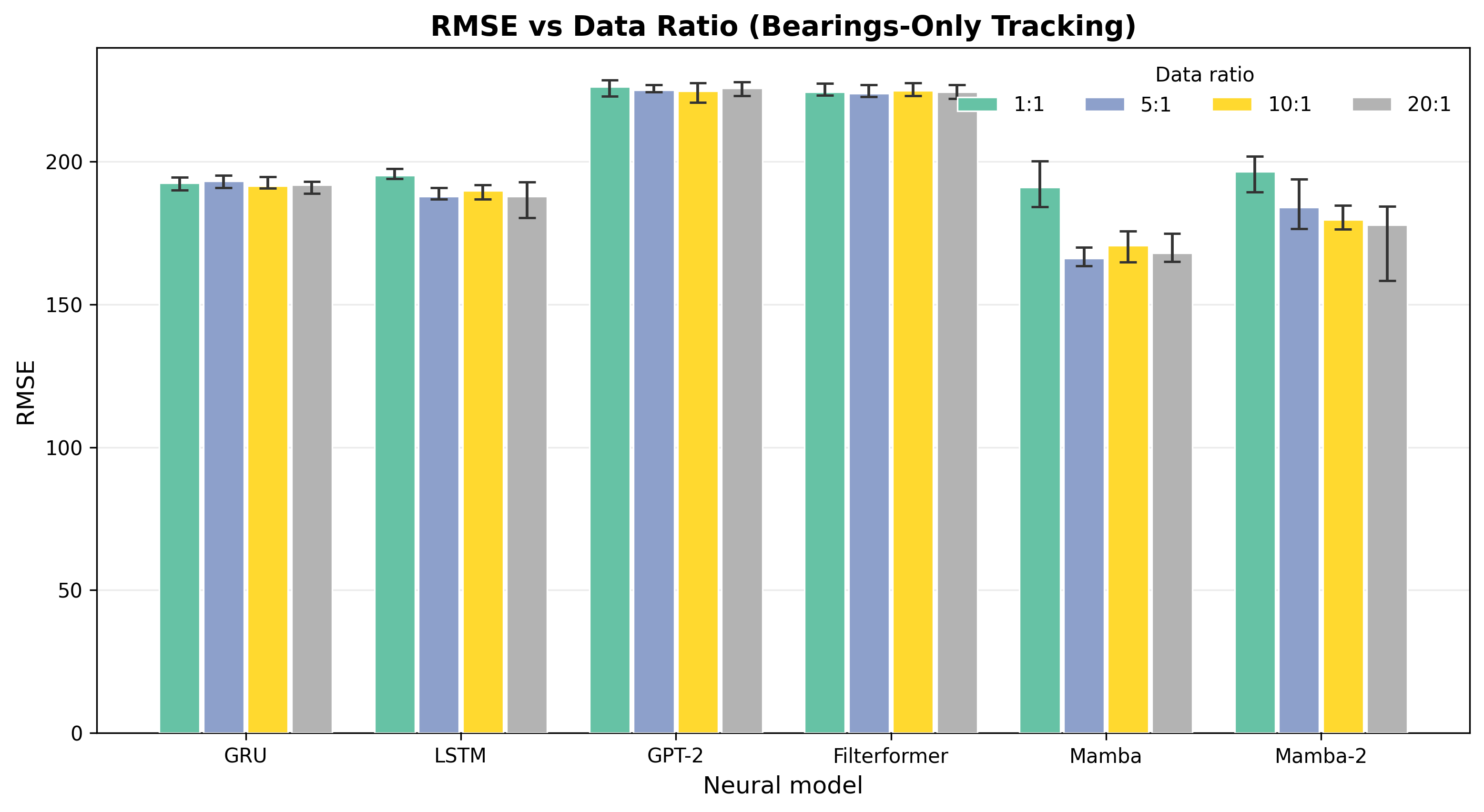}
  \caption{Effect of data-to-parameter ratio on Bearings-only Tracking. Bars show median performance with interquartile-range intervals.}
  \label{fig:data_ratio_bearings}
\end{figure}

\begin{figure}[htbp]
  \centering
  \includegraphics[width=0.92\textwidth]{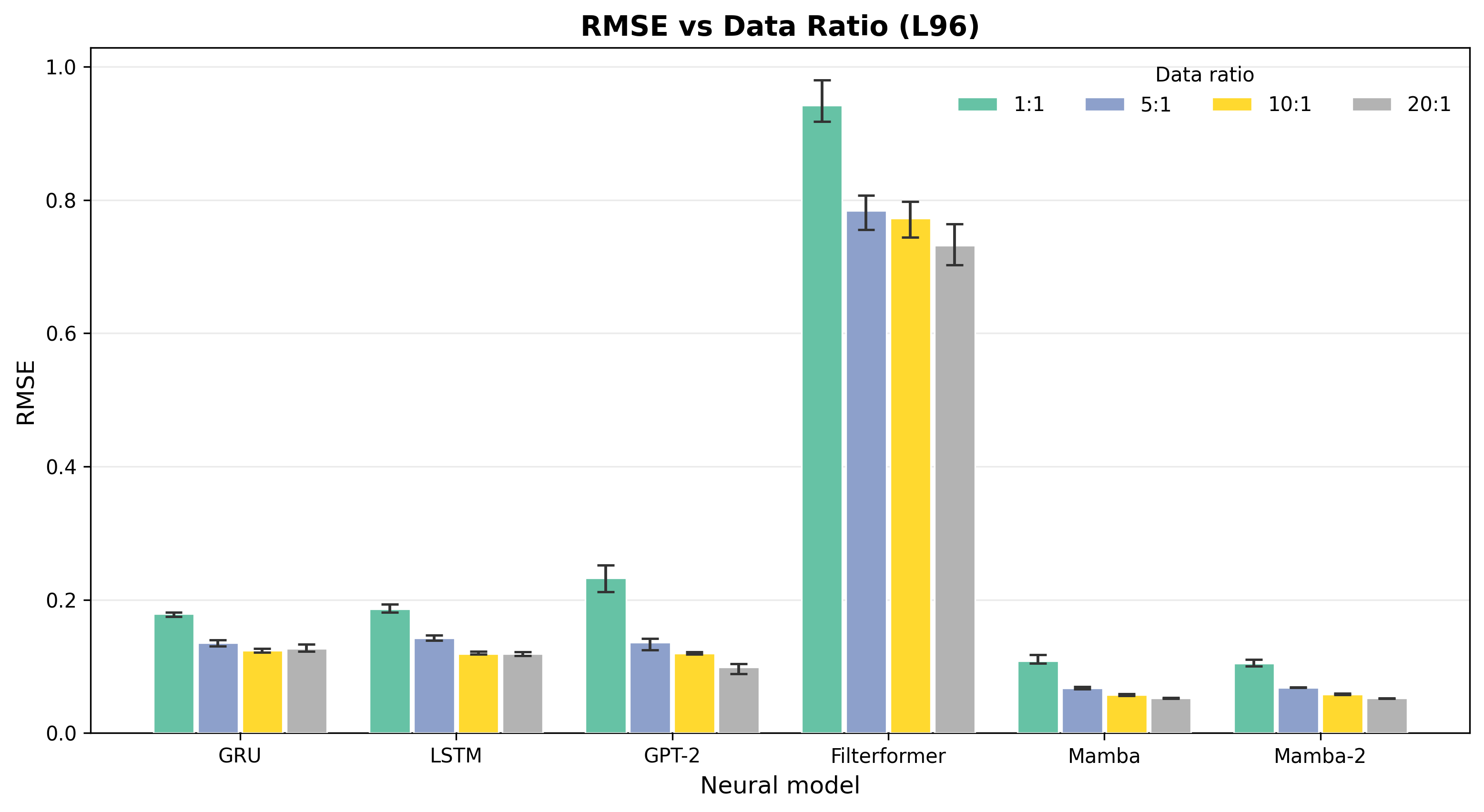}
  \caption{Effect of data-to-parameter ratio on Lorenz 96. Bars show median performance with interquartile-range intervals.}
  \label{fig:data_ratio_l96}
\end{figure}

\begin{figure}[htbp]
  \centering
  \includegraphics[width=0.92\textwidth]{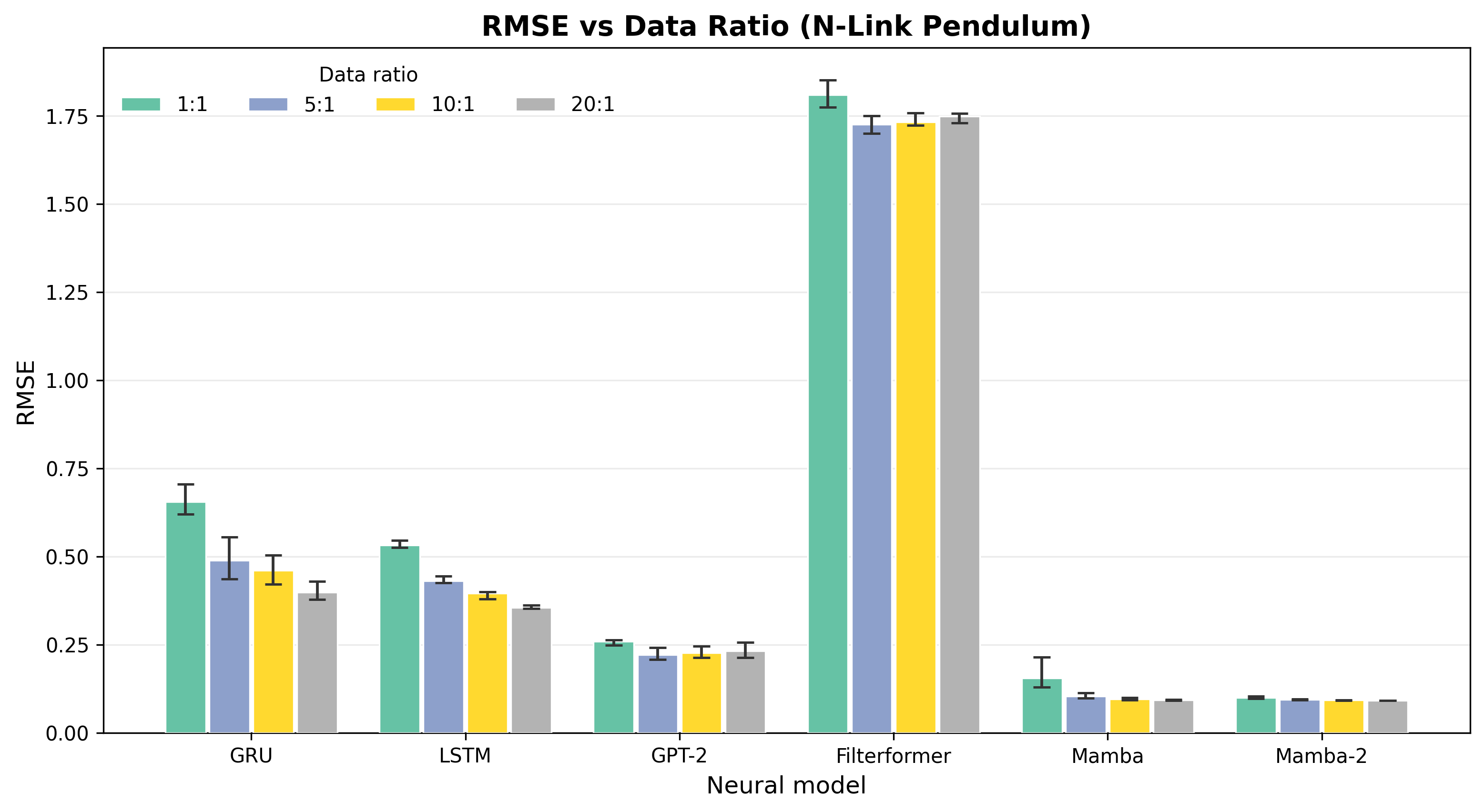}
  \caption{Effect of data-to-parameter ratio on N-link Pendulum. Bars show median performance with interquartile-range intervals.}
  \label{fig:data_ratio_nlink}
\end{figure}

\begin{figure}[htbp]
  \centering
  \includegraphics[width=0.92\textwidth]{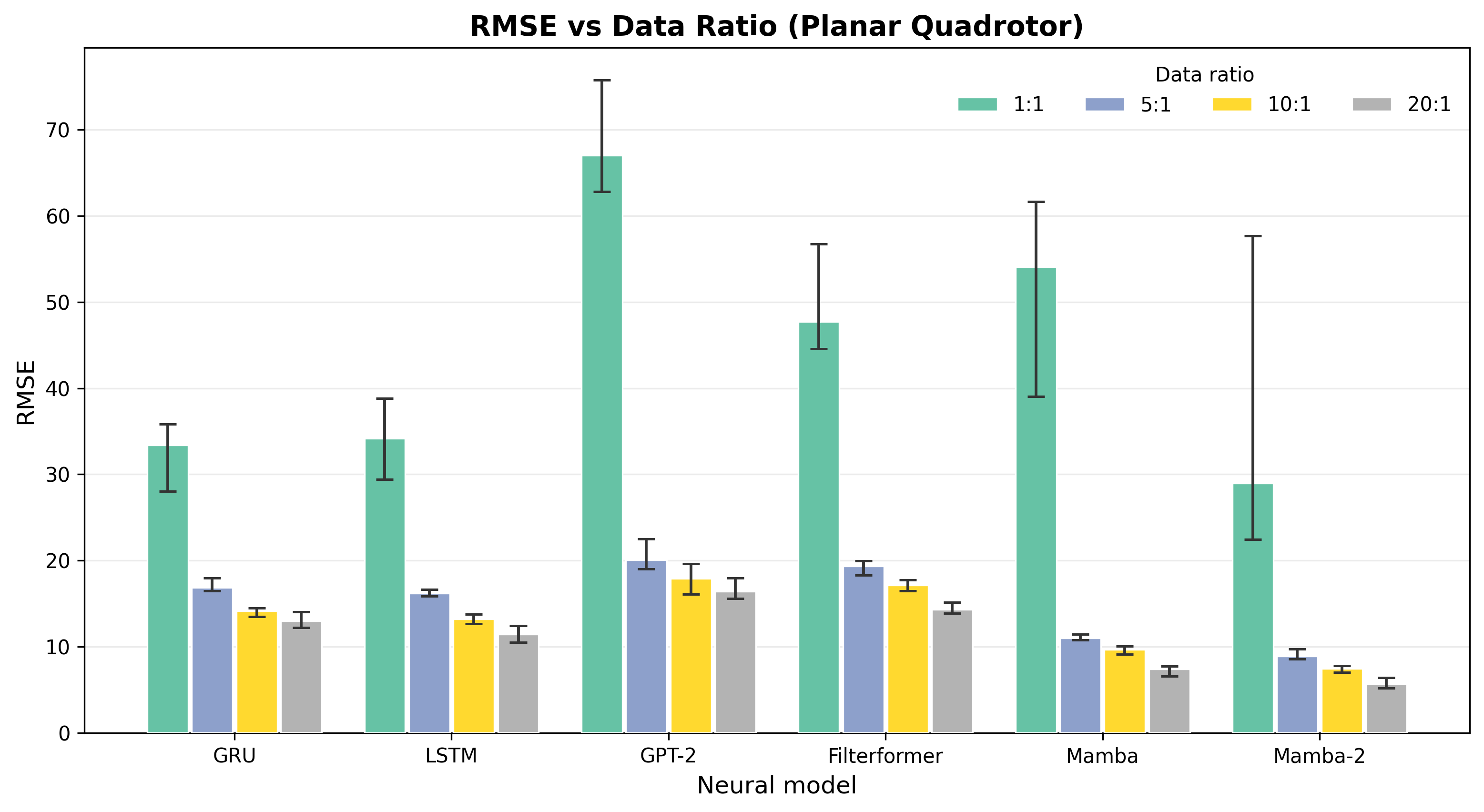}
  \caption{Effect of data-to-parameter ratio on Planar Quadrotor. Bars show median performance with interquartile-range intervals.}
  \label{fig:data_ratio_quadrotor}
\end{figure}

Figures~\ref{fig:data_ratio_ballistic}--\ref{fig:data_ratio_quadrotor} report the effect of the data-to-parameter ratio on neural estimator performance. The default training configuration uses a data-to-parameter ratio of approximately $20{:}1$. Lower-data regimes are evaluated while keeping the model parameter budget fixed. Each bar chart reports the median error for each method and data ratio, with interquartile-range intervals.

\section{Runtime Results}
\label{app:runtime}

Table~\ref{tab:runtime_full_app} reports inference throughput in iterations per second for each method and scenario. Higher values indicate faster inference. CPU results are available for the classical filters and for neural models implemented using standard PyTorch layers. CPU results for Mamba and Mamba-2 are marked as N/A because the official \texttt{mamba\_ssm} implementation used in our experiments relies on CUDA/Triton-backed kernels for the tested forward paths.

\begin{table}[htbp]
  \centering
  \caption{Inference throughput in iterations per second. Higher is better. GPU and CPU results are reported separately.}
  \label{tab:runtime_full_app}
  \footnotesize
  \renewcommand{\arraystretch}{1.05}
  \begin{tabular*}{\textwidth}{@{\extracolsep{\fill}}llccccc}
    \toprule
    Device & Method & Ballistic & Bearings & L96 & N-Link & P. Quad. \\
    \midrule
    \multicolumn{7}{l}{\textit{Classical Filters}} \\
    CPU & EKF & 1029.4 & 7268.1 & 951.2 & 732.4 & 4460.5 \\
    CPU & UKF & 2391.0 & 4135.5 & 1514.8 & 784.0 & 4724.8 \\
    CPU & EnKF & 3789.8 & 10440.5 & 3760.1 & 1202.3 & 9581.5 \\
    CPU & PF & 475.5 & 4538.5 & 1305.1 & 241.0 & 301.7 \\
    GPU & EKF & 294.8 & 1806.1 & 313.6 & 415.9 & 1493.8 \\
    GPU & UKF & 749.0 & 1417.7 & 838.3 & 393.4 & 1436.3 \\
    GPU & EnKF & 1085.4 & 2135.8 & 1295.7 & 468.7 & 3493.3 \\
    GPU & PF & 995.3 & 3165.1 & 1159.5 & 440.8 & 2514.1 \\
    \midrule
    \multicolumn{7}{l}{\textit{Neural Models}} \\
    CPU & GRU & 25290.2 & 26688.1 & 45374.5 & 37867.0 & 22839.1 \\
    CPU & LSTM & 1466.3 & 34264.4 & 129702.2 & 130586.4 & 119974.0 \\
    CPU & GPT-2 & 42194.0 & 8193.6 & 296590.5 & 244805.4 & 263630.5 \\
    CPU & Filterformer & 25776.5 & 14508.1 & 455470.2 & 3232.8 & 1702.1 \\
    CPU & Mamba & N/A & N/A & N/A & N/A & N/A \\
    CPU & Mamba-2 & N/A & N/A & N/A & N/A & N/A \\
    GPU & GRU & 1215110.8 & 1215882.5 & 1216086.6 & 1221712.3 & 913668.9 \\
    GPU & LSTM & 1057379.1 & 1053724.3 & 1060371.1 & 1056281.6 & 821506.2 \\
    GPU & GPT-2 & 434507.5 & 448528.8 & 454934.0 & 454246.0 & 334272.0 \\
    GPU & Filterformer & 3183640.3 & 3109537.4 & 3139871.0 & 3155588.1 & 1443108.3 \\
    GPU & Mamba & 1545250.9 & 1570425.7 & 1577115.2 & 1556686.5 & 632990.2 \\
    GPU & Mamba-2 & 432831.2 & 453866.8 & 457754.7 & 455273.7 & 180400.9 \\
    \bottomrule
  \end{tabular*}
\end{table}


\end{document}